\documentclass[sigconf, 10pt]{acmart}
\usepackage{amsmath,amsfonts}
\usepackage{algorithm, algorithmic}
\usepackage{graphicx}
\graphicspath{ {./figures/} }
\usepackage[utf8]{inputenc}
\usepackage[normalem]{ulem}
\usepackage{soul}
\usepackage{times}
\usepackage{array} 
\usepackage{helvet}
\usepackage{courier}
\usepackage{hyperref} 
\usepackage{booktabs} 
\usepackage{multirow}
\usepackage{colortbl}
\usepackage{subcaption}
\usepackage{wrapfig}
\usepackage{enumitem}
\usepackage[utf8]{inputenc}
\usepackage{multirow}
\usepackage{gensymb}
\usepackage{balance}

\newcommand{\name}{{\em DeepLight }}
\newcommand{\names}{{\em DeepLight}}
\newcommand{\namef}{{\em DeepLight's\ }}
\newcommand{\namet}{DeepLight}

\newcommand{\bfblue}[1]{\textbf{\textcolor{blue}{#1}}}

\setcopyright{none}
\begin{document}

\title[\emph{DeepLight}: Robust \& Unobtrusive Real-time Screen-Camera Communication]{\emph{DeepLight}: Robust \& Unobtrusive Real-time Screen-Camera Communication for Real-World Displays}



\author{Vu Tran}
\affiliation{%
 \institution{Singapore Management University}}

\author{Gihan Jayatilaka}
\affiliation{%
 \institution{University of Peradeniya}}

\author{Ashwin Ashok}
\affiliation{%
 \institution{Georgia State University}}

\author{Archan Misra}
\affiliation{%
 \institution{Singapore Management University}}


\begin{abstract}

The paper introduces a novel, holistic approach for robust Screen-Camera Communication (SCC), where video content on a screen is visually encoded in a human-imperceptible fashion and decoded by a camera capturing images of such screen content. We first show that state-of-the-art SCC techniques have two key limitations for in-the-wild deployment: (a) the decoding accuracy drops rapidly under even modest screen extraction errors from the captured images, and (b) they generate perceptible flickers on common refresh rate screens even with minimal modulation of pixel intensity. To overcome these challenges, we introduce \names, a system that incorporates machine learning (ML) models in the decoding pipeline to achieve humanly-imperceptible, moderately high SCC rates under diverse real-world conditions. \namef key innovation is the design of a Deep Neural Network (DNN) based decoder that collectively decodes all the bits spatially encoded in a display frame, without attempting to precisely isolate the pixels associated with each encoded bit. In addition, \name supports imperceptible encoding by selectively modulating the intensity of only the Blue channel, and provides reasonably accurate screen extraction (IoU values $\ge$ 83\%) by using state-of-the-art object detection DNN pipelines. We show that a fully functional \name system is able to robustly achieve high decoding accuracy (frame error rate < 0.2)  and moderately-high data goodput ($\ge$0.95 Kbps) using a human-held smartphone camera, even over larger screen-camera distances ($\approx$ 2m).

\end{abstract}

\ccsdesc[500]{Human-centered computing~Ubiquitous and mobile computing}
\ccsdesc[500]{Computer systems organization~Embedded and cyber-physical systems}

\keywords{Screen-camera communication, Visible light communication, Imperceptible, Deep neural networks, Perception, Flicker-free}


\maketitle

\section{Introduction}\label{sec:intro}
\begin{figure}[t]
	\centering
	\includegraphics[width=\columnwidth]{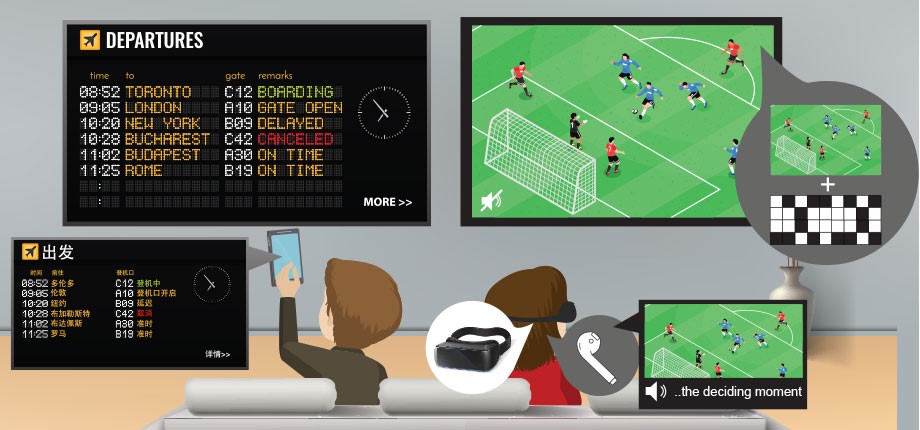}
	\vspace{-0.7cm}
	\caption{An SCC based audio \& text communication use-case scenario in an airline lounge, where SCC is used to encode and transmit a stream of supplementary information accompanying the display screens' content.}
	\vspace{-0.5cm}
	\label{fig:scc-usecase}
\end{figure}

Screen-Camera communication (SCC) is an emerging variant of visible light communication (VLC), where the light source is a  large, individually-controllable, multi-pixel LED screen  and the receiver decodes the visually-encoded information captured by a camera sensor. SCC has evinced strong interest~\cite{woo2012vrcodes, wang2014inframe, wang2015inframepp, li2015real, zhang2019chromacode}, especially as an enabler of ubiquitous computing applications for subliminal machine-machine communication. SCC’s unique challenge is to sustain high throughput while simultaneously ensuring that the visual encoding remains \emph{imperceptible}, when applied to \emph{unregulated} and dynamic video content displayed on regular displays. Unfortunately, while recently-proposed SCC techniques can achieve very high communication rates under controlled conditions at short screen-camera distances, we shall show that these techniques are \textbf{not robust enough} to support diverse real-world  artefacts, such as (a) longer (1.5-2m) screen-camera distances and non-frontal viewing angles, (b) lack of a-priori knowledge of precise screen coordinates, and (c) continuous perturbation of camera views due to human movement.

\begin{figure*}
\begin{minipage}{.48\textwidth}
	\centering
	\includegraphics[width=\columnwidth]{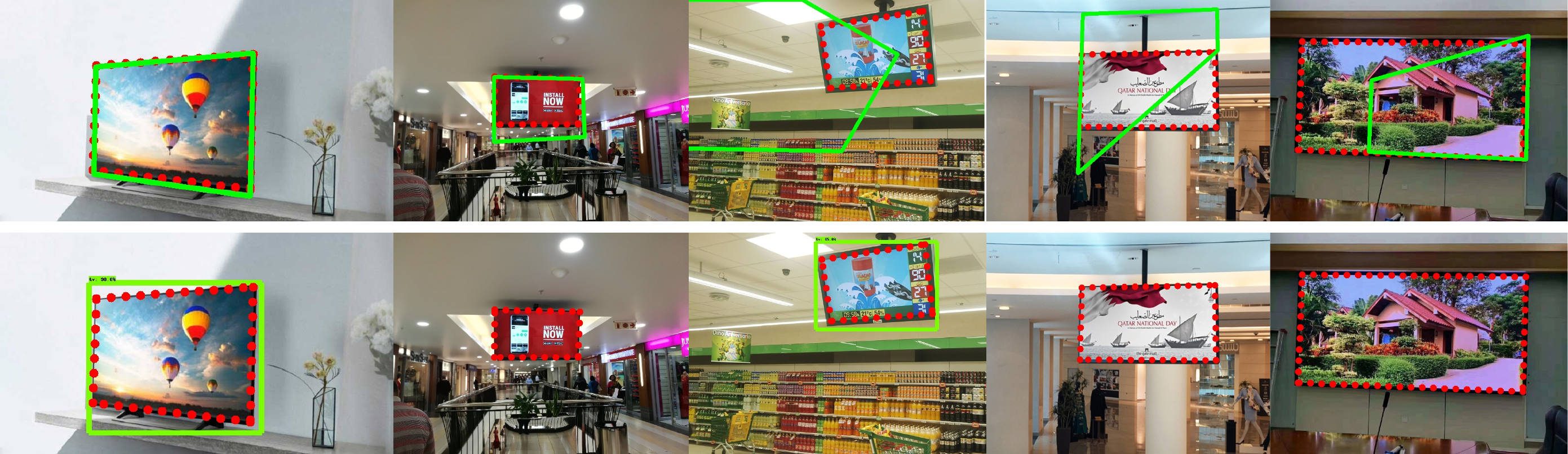}
	\vspace{-7mm}
	\caption{Screen extraction performance of \emph{(1st row)}: Edge analysis algorithm, and \emph{(2nd row)}: SSD-MobileNet, with photos of representative displays. The ground-truth borders are drawn in ``dotted red'' lines while the detected borders/bounding boxes are drawn in ``solid green'' lines.}
	\vspace{-0.3cm}
	\label{fig:screenshots}
\end{minipage}
\hfill
\begin{minipage}{.48\textwidth}
    \centering
	\includegraphics[height=1.2in,width=\columnwidth]{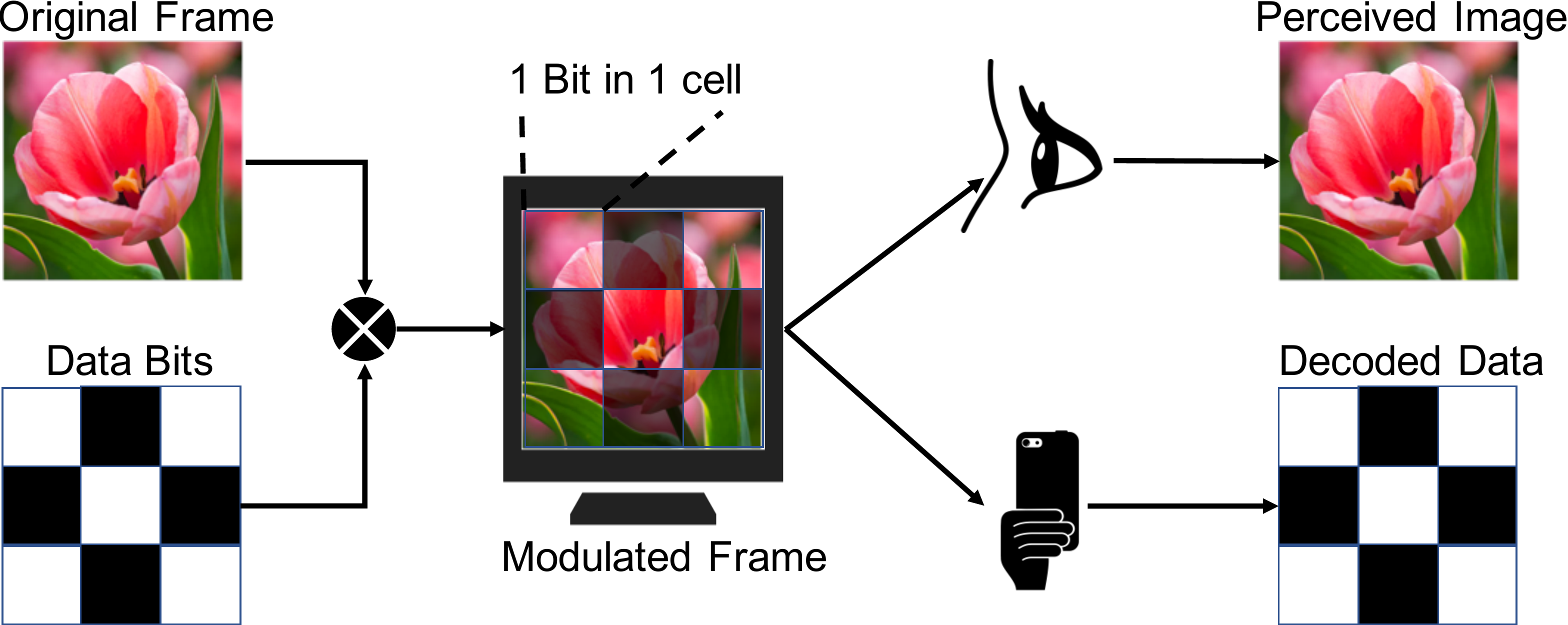}
	\vspace{-5mm}
	\caption{SCC Fundamentals. Bits are spatially encoded on the display frame for decoding by a camera, while staying  imperceptible to human eye.}
	\vspace{-0.3cm}
	\label{fig:scc-basic}
\end{minipage}
\end{figure*}

In this paper, we describe \names, a novel SCC system that \emph{employs machine learning (ML) based inferencing techniques in the decoding pipeline to achieve a degree of robustness, to the artefacts mentioned above, that has hitherto been unattainable}. 
To our knowledge, \name is the first system to demonstrate \emph{low-flicker} SCC for real-world scenarios involving \emph{commodity} (screen refresh rates of 60Hz) public display screens and handheld smartphone cameras, where: (a) an individual user points the camera sensor towards the screen from  different viewing angles and distances, (b) the screen content can include both still images and dynamic video streams, and (c) the screens are located in environments with different ambient lighting levels and backgrounds (e.g., fast moving crowds vs. walls) and have attributes that are \emph{a priori} unknown to the mobile camera. 

Figure~\ref{fig:scc-usecase} illustrates a couple of exemplar \names-based ubiquitous computing applications (including one that we have built and deployed). Here, SCC is used to visually and subliminally encode the following supplementary information: (a) User U1 watches the sports feed with a see-through head-mounted, camera-equipped display: the wearable \name software is able to decode the SCC-encoded \emph{audio} commentary, in real time, and then output this via  Bluetooth-enabled earphones; (b) 
user U2 points a camera-equipped smartphone at the flight information display: the mobile-embedded \name  decodes the \emph{textual} commentary(encoded in multiple languages via SCC) and overlays the information in U2’s native language.
 
\vspace{2mm}\noindent \textbf{Key Technical Challenges \& Innovations of \names\footnote{Source code \& demo video available at:\\ \url{https://larc-cmu-smu.github.io/deeplight/}}:} Our work provides the following breakthroughs to address several key challenges:
\renewcommand\labelitemi{\tiny$\bullet$}
\begin{itemize}[leftmargin=*]
\item \emph{A CNN-based Collective-Bit-Decoder to Make Decoding Robust to Imprecise Screen Detection:} Current SCC decoding techniques operate by effectively \emph{explicitly} dividing the estimated screen, in a grid-like fashion, into a set of constituent cells and then independently inferring the encoded bit of each such cell. Consequently, these SCC techniques suffer a dramatic jump in bit-error-rate (BER) even under relatively modest screen extraction errors--e.g., the BER of HiLight increases by 12\% even if the detected screen is offset by only 10\% of a cell. To support robust decoding of transmitted bits without needing a perfectly-extracted screen, \name uses a novel \emph{Collective Bit Decoder}, a convolutional neural network (CNN) model that \emph{collectively} infers the entire set of bits in a display frame (instead of trying to explicitly isolate the pixels for each individual bit). \namef use of CNN-based models for decoding represents a significant departure from prior signal processing-based approaches, and is particularly effective in extracting the bit information (encoded as intensity changes in the blue channel, as explained shortly) from a multi-scale analysis of the image content. Through extensive experiments, we show that this CNN is able to achieve frame error rates (FER) $\le 1.7\%$ (an improvement of almost 15\% over prior techniques), even when the IoU (Intersection over Union) of screen detection is modest ($IoU_{screen}=96\%, IoU_{cell}=68\%$).

\vspace{2mm}
\item \emph{Ensure Accurate Real-world Screen Detection \& SCC Operation without Needing Explicit Screen Markers:}
Prior work typically utilizes either static cameras (with pre-calibrated screen boundaries) or special markers to delineate the screen edges. Such precise extraction of screen boundaries is not practical, even with state-of-the-art DNN models, for real-world displays. For example, Figure~\ref{fig:screenshots} shows a representative set of real-world images containing display screens in public spaces, together with (1) the detected screen border estimated by a common edge analysis algorithm (Canny \& Hough transform), and (2) bounding boxes for `display' objects identified by the popular SSD~\cite{liu2016ssd} object detector. We experimentally observe that the edge analysis algorithm 
fails to extract screens when the content or the background includes many edges. Consequently, the average IOU is considerably low (38\%) over a set of 266 representative images of TV screens in public spaces. Similarly, SSD fails to detect screens where the content and the background contain large objects, and thus can correctly detect only 73\% of screens in the same test set. Though SSD is not designed to accurately estimate the border of objects (only bounding box), it still achieves, on average, an IOU of 45\% (higher than the edge analysis algorithm). This suggests that a properly designed deep-neural network may support sufficiently accurate screen extractions. To improve screen extraction accuracy, we develop a DNN-based pipeline that can extract the coordinates of the display screen,  without assuming any external markers or a-priori knowledge, with sufficiently high accuracy (average IoU=93\% in indoor environments) to permit robust decoding by our Collective Bit Decoder. Moreover, this pipeline is lightweight enough to achieve screen extraction latency of < 40 msecs/frame on an iPhone 11Pro device. 

\item \emph{Overcome High Flicker Perception at Common Frame Rates:} Most of the proposed techniques (e.g.,~\cite{wang2015inframepp,nguyen2016high, li2015real, zhang2019chromacode}) involve modulation of the pixel intensity \emph{along all 3 (RGB) channels}. While such RGB modulation is often imperceptible at very high display frame rates (e.g., in excess of 120 fps) used by specialized displays, even minimal intensity changes ($\pm1$ out of 255) become noticeable at the typically lower refresh rates (30-60 fps) commonly used in public displays. To tackle this problem, \name encodes the bits via intensity changes ($\Delta$ units) only in the \emph{Blue}(`B') channel of display content. Through experimental studies, with 17 users and viewing distances of 1-2 meters, we show that this modulation scheme causes significantly less flickers (mean opinion scores (MOS) is $\approx20\%$ higher) even when the intensity is modulated by a larger amount ($\Delta=3$). 
\end{itemize}

Besides demonstrating these core component technologies, we develop an end-to-end functional prototype of \name (deployed as a self-contained iOS App that implements the subtitle overlay application described before), which additionally uses an appropriate error-correcting source coding scheme to support high data goodput across diverse real-world settings. Via extensive empirical experimentation, we show that \name can significantly outperform state-of-the-art competitive baselines, achieving both high data goodput (1-1.2 Kbps) and low flicker perception (MOS values of 4 and higher) for larger screen-camera distances (up to 2 meters) and  diverse viewing angles ($0- \pm 30^\circ$).

\section{Background and Related Work}\label{sec:related}
The conceptual idea of SCC is illustrated in Figure \ref{fig:scc-basic}. At the display screen or {\em transmitter} end, the display content image (or frames of video) pixels are logically divided into a rectangular grid, where \emph{each cell in the grid represents at least one encoded bit}. The data bits are modulated via subtle changes in corresponding pixels by an amount of $\Delta$. A practical SCC system must ensure that the camera (i.e., the {\em receiver} device) is able to decode this original embedded information, even under distortions such as non-linear camera optics, effects of ambient lighting and motion artifacts (caused by the mobility of the camera-embedded device). While decoding accuracy can be enhanced by using a larger $\Delta$, this results in the human eyes perceiving a flickering chessboard-like pattern. Thus, the central challenge in SCC is to achieve higher throughput  while avoiding perceptible flickers. Prior work has focused primarily on designing reliable and efficient (a) encoding mechanisms to \emph{improve throughput}~\cite{wang2015inframepp, shi2015reading, zhang2019chromacode}, and/or (b) embedding mechanisms to \emph{reduce flicker perception}~\cite{li2015real, nguyen2016high, woo2012vrcodes}. Broadly, the encoder schemes follow two approaches:
(a) Manchester encoding, or (b) Frequency encoding.

\noindent{\bf Manchester encoding based techniques.}
The Manchester encoding technique modulates a single bit over two consecutive frames by adjusting the pixel intensity with an amount of $\pm\Delta$. This approach leverages the fact that human eyes average out content changes at rates beyond $\approx$ 50Hz~\cite{kelly1972flicker, simonson1952flicker, brindley1966flicker, wilkins2010led}; accordingly, if the screen refresh rate $R_S$ is greater than 100 Hz, the flickers are perceptually canceled out. Therefore, Manchester encoding is visually imperceptible only at high display rates (unless $\Delta$ is very small) and is ineffective for commonly-used displays with $R_S=$ 50/60Hz. VRCode~\cite{woo2012vrcodes} applies Manchester coding on color channel (in CIE color space), and supports only invariant data (e.g, QR Code). In contrast, Inframe++~\cite{wang2015inframepp} supports time-variant data and visual content by encoding data on brightness channel using a novel spatial pattern. However, the variable data transmission causes flickers as it creates additional harmonics. TextureCode~\cite{nguyen2016high} modulates only the parts of the frame with higher texture to reduce flickers. Chromacode~\cite{zhang2019chromacode} uses LAB~\cite{cielab} color space, and applies adaptive texture encoding for full frame modulation. All these approaches require precise screen extraction (which is unrealistic in practice), as each bit is decoded individually, based on a predefined, grid-based partitioning of this captured screen.
    
\noindent {\bf Frequency Modulation based techniques.} HiLight~\cite{li2015real} encodes a bit ``1" and bit ``0" as variations at 30Hz and 20Hz over 16 frames (thereby reducing the overall throughput) respectively. HiLight adjusts the screen's brightness (commonly referred to as $\alpha$ channel in graphics) using low $\alpha$ to suppress flicker. Though HiLight does not use special patterns to estimate screen border, it requires turning the screen OFF/ON for precise screen extraction.

\noindent{\bf Barcode Communication.}
Extensive prior work~\cite{ohbuchi2004barcode,liu2008recognition} has explored the process of smartphone-based 2D barcode or QR code scanning, which may be viewed as a rudimentary form of SCC. Several approaches (\cite{pixnet2010, hu2013lightsync, du2016softlight}) have investigated the transmission of \emph{time-varying QR-Code} data as a form of high bit-rate SCC, with MegaLight~\cite{zhan2019megalight} using a Random Forest-based ML model to support robust decoding. However, all of these techniques consider the QR code as the \emph{primary} display content and do not attempt to make the transmission imperceptible to human viewers.

\vspace{1.5mm}\noindent{\bf Deep Learning based Watermarking \& Steganography.}
The idea of `hiding' information into image content is borrowed from the well-known concept of watermarking or steganography~\cite{provos2003hide, cox2007digital}. Most classical steganographic techniques, which either modify the least significant bits (LSBs) in pixels~\cite{morkel2005overview} or use DCT~\cite{barni1998dct}
, have very low data rates (bits/frame). While deep learning has recently been explored as a technique for enhancing capacity~\cite{baluja2017hiding}, the techniques fundamentally assume a lossless transfer of the encoded image to the receiver. The recent work by Wengrowski and Dana~\cite{wengrowski2019light} is the first to utilize a DNN-based framework for screen-camera steganography. 
However, this work focuses on photographic steganography (where the encoded image must be extracted precisely) and requires the learning of a \emph{per-display} transfer model (that precludes seamless in-the-wild adoption).

\section{Limitations in Existing Methods}\label{sec:limitations}
\begin{figure*}
\begin{minipage}{.49\textwidth}
 \centering
 \begin{tabular}{cc}
         \includegraphics[width=0.50\textwidth]{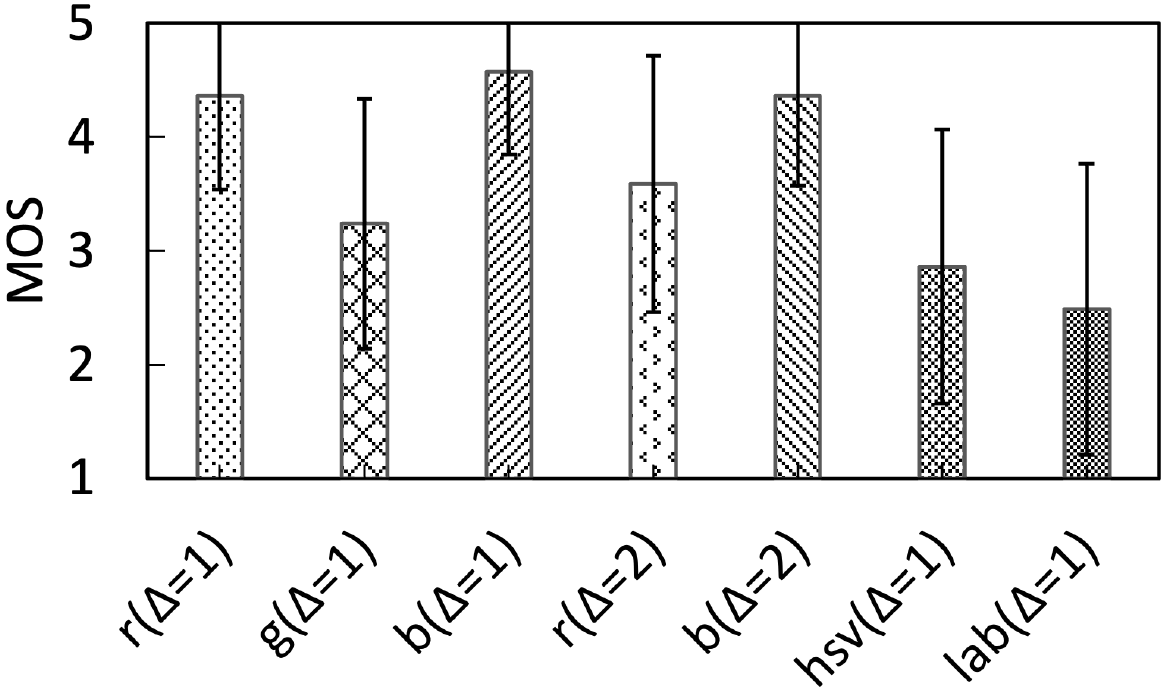} & 
         \includegraphics[width=0.42\textwidth]{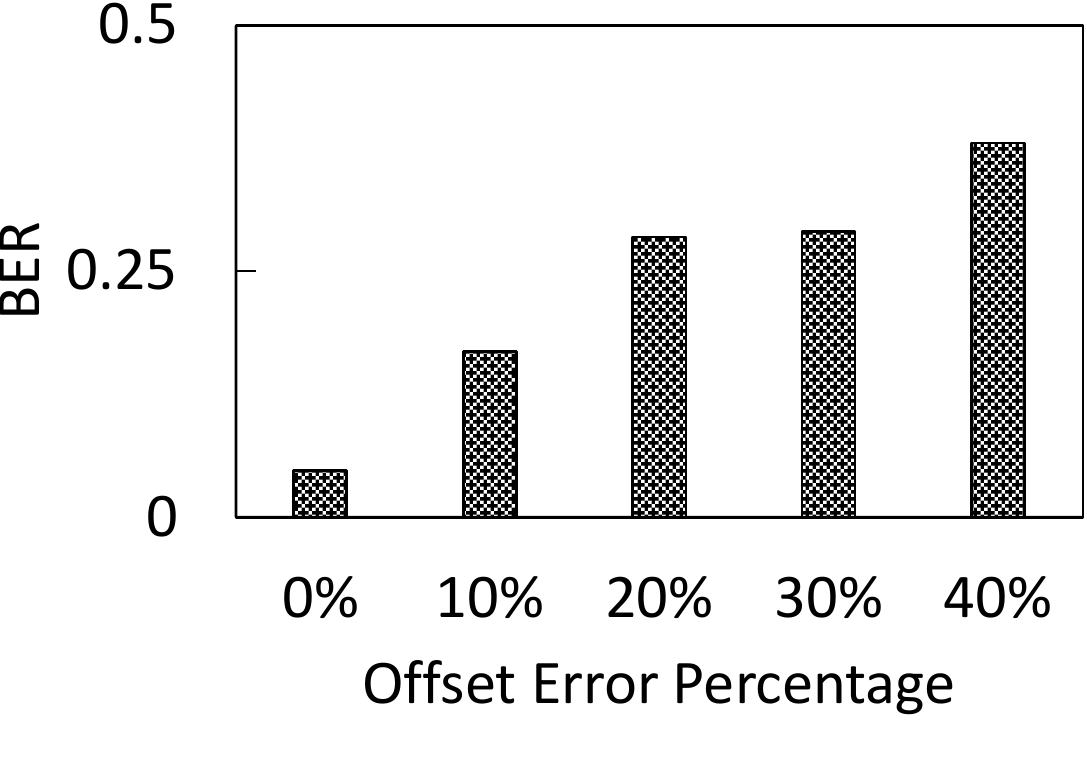} \\
  \end{tabular}
  \vspace{-5mm}
   \caption{(Left) Flicker score of intensity-modulated video clips (distance= 1 meter, $\Delta=\{1,2\}$).~(Right) \emph{HiLight} BER  under varying screen extraction error.}
   \vspace{-0.2cm}
   \label{fig:limitationsall}
\end{minipage}%
\hfill
\begin{minipage}{.49\textwidth}
 \centering
 \includegraphics[width=\columnwidth]{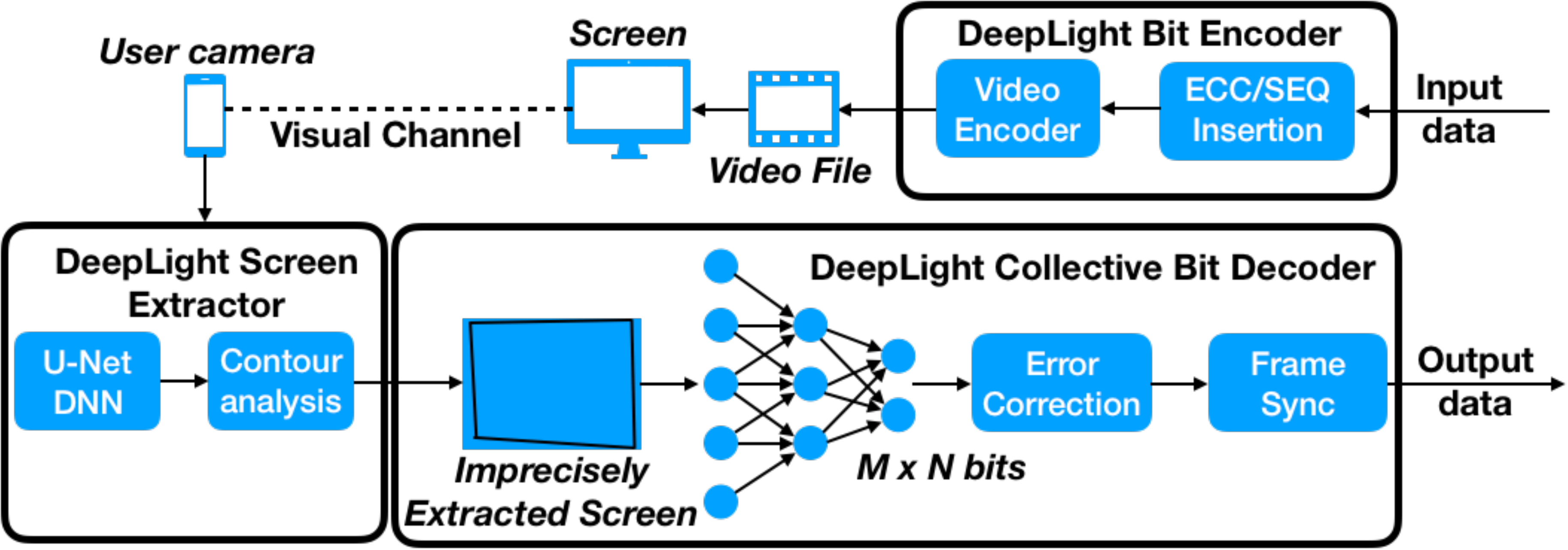}
 \caption{\names: Overall System Architecture}
 \vspace{-0.2cm}
 \label{fig:overview}
\end{minipage}%
\end{figure*}

Before presenting the design of \names, we shall experimentally quantify two key limitations of prior SCC methods, focusing on, (a) human {\em perception of flickers} emanating from the screens due to embedded modulation, and (b) vulnerability of the system performance to {\em imprecise screen extraction}.

\vspace{-0.1in}
\subsection{High Flicker Perceptibility} 

It is essential for SCC is to preserve the visual quality of modulated video frames. 
Prior works rely on high frequency (120FPS or higher) displays to suppress flickers caused by modulation schemes \cite{wang2015inframepp, shi2015reading, nguyen2016high, zhang2019chromacode}, whereas commercial displays in public spaces usually have 30-60 FPS refresh rate. It turns out that even minimal modulation ($\Delta=\pm1$) causes significant flickers at 60FPS, making current modulation approaches, which change all 3 channels (via brightness channel~\cite{wang2015inframepp, shi2015reading, nguyen2016high, zhang2019chromacode} or alpha-blend~\cite{li2015real}), inapplicable to common displays. Fortunately, humans are known to be relatively less sensitive to Blue light (compared to Red \& Green)~\cite{webster1996human, brindley1966flicker}, due to the presence of fewer Blue-sensitive photo-receptors in our eyes~\cite{roorda1999arrangement}. In this paper, we show that, by modulating the data on Blue channel only, our system achieves the highest (and considerably higher) mean opinion score (MOS) compared to prior works, while achieving a higher goodput at practical viewing distances ($\geq 1m$). 

To validate this effect in SCC, we conduct a pilot study on flicker perception with modulations on different channels. Figure \ref{fig:limitationsall} (left) plots the Mean Opinion Score (MOS) and its variation, across 12 users, who rated the perceived quality of a total of 42 video clips (each clip is 10 secs long), displayed on a 60FPS 25"  monitor, and viewed from $d=$1 meter. The videos were encoded using the widely-used Manchester coding technique~\cite{nguyen2016high, wang2015inframepp, zhang2019chromacode}, with the barest possible pixel modulation amplitude ($\Delta=\pm1$, out of a total range of 255). Each user was exposed to different treatments: (a) where the modulation is performed on brightness channel (under HSV or LAB color spaces) or a single color (R, G or B) channel, and (b) where the modulation magnitude is higher ($\Delta=\pm2$). We see that, even with the least possible modulation amplitude ($\Delta=\pm1$) on brightness channel or Green channel, the MOS score  is significantly lower ($\le3.3$). Though Red channel achieves fairly high MOS (avg $= 4.36$) with low modulation amplitude ($\Delta=\pm1$), it performs much worse (avg $= 3.58$) with higher $\Delta$. However, the MOS is consistently high if only the Blue channel is modulated (avg $ = 4.57$ and $4.36$ with $\Delta=\pm1$ and $\pm2$ respectively), validating the physiological fact. This suggests that performing Blue-channel-based modulation \emph{may} considerably enhance the imperceptibility of SCC for larger displays, at common display rates.

\vspace{-0.1in}
\subsection{Vulnerability to Imprecise Screen Extraction}
\label{subsec:screenerror}
Prior works assume perfect screen extraction achieved using either explicit visible locators (e.g. InFrame++~\cite{wang2015inframepp}, QRCode~\cite{ohbuchi2004barcode}) or adding an explicit set of black and white strips next to the screen border (e.g., ImplicitCode~\cite{shi2015reading}, Chromacode~\cite{zhang2019chromacode}). Neither of these is compatible with our goal of in-the-wild SCC using unmodified displays. Such precise screen extraction is necessary because each bit is decoded \emph{independently} using an explicit partitioning of the rectangular screen into a $M\times N$ grid. Accordingly, a lack of precise extraction causes a specific cell to either include superfluous pixels (from neighboring cells) or excluding constituent pixels (which are incorrectly mapped to a neighboring cell). Unfortunately, even practically ``precise'' screen extractions may easily include a deviation. Even if this deviation is small compared with screen size, it may be significant compared with cell size. For example, the top left image in Figure~\ref{fig:screenshots} clearly shows an offset between the estimated lower edge (using Canny \& Hough transform) and the ground-truth (even if the screen extraction is considered ``precise''), which affects the cells near the lower edge severely. However, as illustrated in Figure~\ref{fig:screenshots}, accurate estimation of an active screen's borders is not trivial under real-world artifacts. The fact that a display shows dynamic contents, with a plethora of textures and colors, makes ``precise'' estimation of the screen border impractical. Indeed, past SCC prototypes have used custom screen locators~\cite{wang2015inframepp, shi2015reading, zhang2019chromacode} or utilized out-of-band techniques to tackle this challenge. As an example of these issues, HiLight~ \cite{li2015real} documented the significant loss of accuracy for viewing distances larger than 1 meter, due to the higher screen extraction error. To tackle this, HiLight's edge detection mechanism (published code in~\cite{hilightcode}) explicitly requires the user to turn the screen OFF/ON at the beginning of each SCC session, an approach that is untenable for public venues.

To quantify this issue,  Figure~\ref{fig:limitationsall} (right) plots our empirically observed bit-error-rate (BER) values with HiLight (viewing distance $d=0.7m$ meter), for varying degree of synthetically generated extraction errors (expressed as a percentage of the size of a single cell of a $10 \times 10$ grid). We study HiLight as it is the only prior work whose code we can obtain publicly, and it applies the same approach of explicit grid splitting used in other works. We see that the decoding accuracy drops significantly, even when the extraction error is as low as 10\% (a paltry error of $19 \times 11$ pixels in our HD display). At 20\% offset error, the BER is $\approx 29$\%, an increase of $\approx$25\% in BER compared with the perfect screen extraction. We shall demonstrate (Section~\ref{subsec:screenextractor}) how the use of ML pipelines in \name allows us to overcome these limitations.

\subsection{\textbf{\namet~Design Goals}}
To achieve the operational vision of a robust and practical SCC system, \name should ideally achieve the following objectives:
\begin{itemize}[wide, labelindent = 0pt, topsep = 1ex]
\item \emph{Support Relatively-High SCC Data Rates:} \name use cases extend well beyond the transmission of relatively static content (e.g., QR-codes), and thus require at least modestly-high data goodput. In particular, with advanced encoding~\cite{schroeder1985code} human speech transmission (for user U1 in Figure~\ref{fig:scc-usecase} earlier) can be achieved with bit rates of 1-2 Kbps; whereas for transmission of subtitles (for user U2 Figure~\ref{fig:scc-usecase} earlier), the maximum acceptable rate is 200 WPM (word-per-minute), effectively translating into  a target rate of $\sim 0.2$ Kbps (assuming the support of 8  languages).  To support such SCC applications, {\bf \name should support data rates of at least $0.5$ Kbps, and ideally $2+$ Kbps.}

\item \emph{Support Variations in Viewing Distances, Angles \& Backgrounds:} In public spaces, viewers may use their mobile devices to decode the SCC display from a variety of viewing distances and viewing angles. For robust and practical SCC, {\bf \name should ideally support reasonably-high decoding goodput (across a variety of spaces such as office cabins, shopping malls and airports) for viewing distances of up to $1.5-2$ meters and viewing angles of up to 60$^\circ$}. 

\item \emph{Work across a Range of Static \& Dynamic Screen Content:} 
{\bf \name should ensure that its visual encoding remains imperceptible across real-world display content}, ranging from dynamic (e.g., the live feed of a sporting event, where the scene changes once every 10-20 ms) to relatively static (e.g., the airport information feed is likely to refresh only once every 5-10 seconds). 

\item \emph{Work with Imprecise Screen Capture:}  Based on the limitations presented above, {\bf \name must be capable of delineating the boundary/contour of diversely-shaped display screens (within a captured image) with reasonable accuracy, and subsequently decoding the embedded content under imperfect screen extraction}.
\end{itemize}

\section{\name System \&  Components}\label{sec:system}

\begin{figure}
    \begin{center}
        
    \includegraphics[width=\columnwidth]{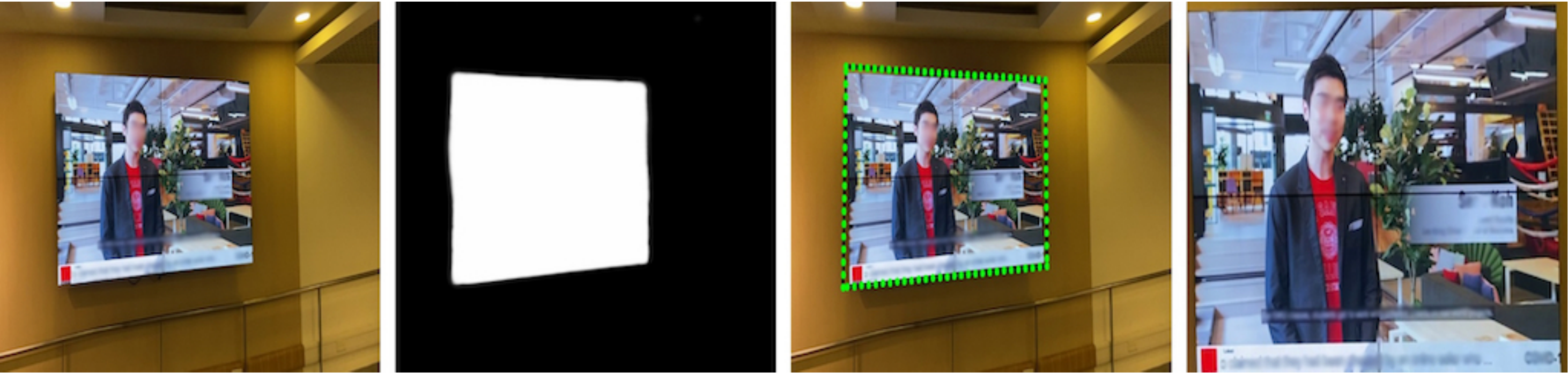}
    \vspace{-0.7cm}
    \caption{Four stages of screen extractor. L to R: Input, Unet output, Post-segmentation Screen (dotted green bounding box), Transformed Screen}
    \vspace{-0.5cm}
	\label{fig:3stage-extractor}
	\end{center}
\end{figure}

\begin{figure*}[t]
\begin{minipage}{.6\textwidth}
 \centering
 \includegraphics[width=\linewidth]{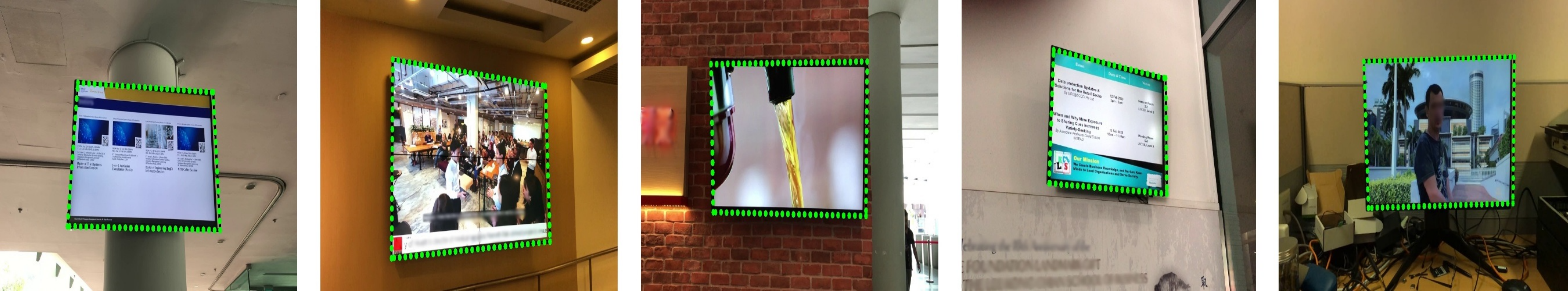}
 \vspace{-6mm}
   \caption{Applying Screen Extractor to different screens \& backgrounds, spanning office and public spaces. Dotted green bounding box is the estimated screen.}
   \vspace{-0.3cm}
 \label{fig:different_screens}
\end{minipage}%
\hfill
\begin{minipage}{.38\textwidth}
\vspace{-0.25cm}
\centering
	\includegraphics[width=\linewidth]{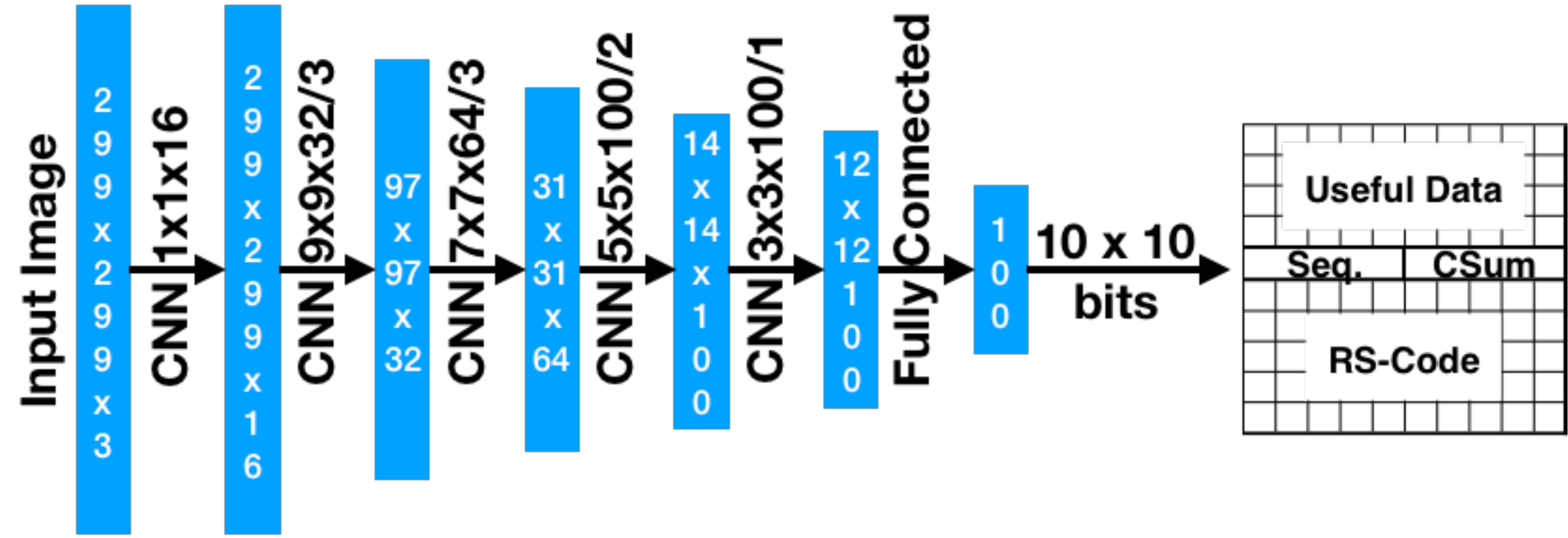}
	\vspace{-7mm}
	\caption{\name CNN Decoder \& Structure of data frame.}
	\vspace{-0.3cm}
	\label{fig:cnn_model_features}
\end{minipage}%
\end{figure*}

To tackle the limitations enumerated above, we propose a new SCC system, called \names, that is based on three key ideas: (a) on the encoding side, the information bits are encoded via modest intensity changes of only the display content's Blue channel, (b) the use of a state-of-the-art DNN on the decoder to extract a display screen \emph{fairly accurately} from a captured image, under a variety of ambient context and (c) the design of a new DNN model that provides holistic/joint decoding of multiple bits in a display image, even when the screen extraction is imperfect. 

Figure~\ref{fig:overview} illustrates the high-level architecture of \names. On the encoder side, data bits are packed and embedded into video frames before being shown on a screen. A camera sensor (e.g., belonging to a user's smartphone) then captures images at $f_{C}$ FPS, with the display screen assumed to be visible (but at an arbitrary angle and position) within each captured image. On the decoder side, an initial enhanced DNN-based \emph{Screen Extractor} component operates on each image to isolate and extract the pixels estimated to be part of the \names-encoded screen.  A series of $L$ such consecutively ``extracted screens'' is then forwarded to the \emph{Bit Decoder} (this represents the most innovative aspect of \names), which uses another DNN to \emph{collectively} reconstruct the transmitted bits. 

\vspace{-0.1in}
\subsection{Bit Encoder}

To embed data into video frames, \name bit encoder encodes the information data bits by mapping a set of bits to a $M \times N$ rectangular grid, where each cell corresponds to a block of $B_{r} \times B_{c}$ pixels of the image (video frame) to be encoded. In each cell, the bit encoder applies a Manchester coding strategy by performing opposing intensity modulation of the \emph{Blue} channel of each pixel in the cell by a value of $\pm \Delta$ across two consecutive (F+, F-) frames. By extensive trial and error with different $\Delta$ choices (detailed in SectionSection~\ref{sec:evaluation}), we find $\Delta$ values of 2 or 3 to be reasonable, representing a compromise between our desire for lower decoding error (high $\Delta$) and visual imperceptibility (low $\Delta$). These encoded image frames are then displayed on the  screen at its original operational frame rate $F_d$.

\vspace{-0.1in}
\subsection{Screen Extractor} \label{subsec:screenextractor}

At the decoder side, the first step is to localize the \name screen in a camera frame. We first evaluate how well the classical edge analysis approaches support screen extraction. Though many techniques have been proposed to detect rectangular objects (similar to a screen) such as license plates~\cite{chang2004automatic} \& QR-Codes~\cite{ohbuchi2004barcode, liu2008recognition}, ``precise'' screen extraction is tricky for two reasons: (1) there may be parts of the content next to the screen border that have similar color with the screen border; (2) the screen contents and the background usually contain lots of textures. We experimentally observe that classical edge analysis approaches work well with uniformly colored screens, but perform quite poorly (examples shown in Figure~\ref{fig:screenshots}) on our corpus, where screens have high-texture content or occupy a small portion of the overall image.

To locate and extract the screen in a camera frame, \name borrows from state-of-the-art DNN-based object segmentation and detection techniques to isolate screens under a wide variety of settings. The Screen Extractor consists of three cascaded stages (illustrated in Figure~\ref{fig:3stage-extractor}), with additional smoothing and thresholding operations to  improve the noise tolerance: 

\vspace{1mm}\noindent{\bf (Step 1)} A DNN-based pixelwise segmentation pipeline, inspired by Unet architecture \cite{unet2015}, that first assigns a probability for every pixel belonging to a screen. 
This 2D probability distribution is dilated (smoothened) by a uniform kernel and thresholded to assign binary labels (`screen' vs. `no-screen') to each pixel.

\vspace{1mm}\noindent{\bf (Step 2)} The segmented image is then passed through a contour-based screen localizer~\cite{szeliskibook} that ensures that the identified set of `screen' pixels adhere to certain natural shape constraints (e.g., the screen must be a quadrilateral).

\vspace{1mm}\noindent{\bf (Step 3)} Finally, the set of pixels representing the screen is run through a linear-interpolation based perspective transform, which effectively warps the oblique view of the screen (when the screen is captured at non-perpendicular viewing angles) to a normalized `full-frontal view'.

This DNN is trained on a dataset (illustrative examples, which include the bounding boxes estimated by our technique, are provided in Figure~ \ref{fig:different_screens}) consisting of multiple screen types, backgrounds (indoor and outdoor) and camera setups (handheld and tripod). We used multiple techniques to curate a comprehensive training set of representative screen images: (a) using Web search to retrieve over 387 relevant images of public displays from locations such as malls, train stations; (b) personally visiting over 20 real-world locations across  different geographies to collect 35 images of digital screens. The images are captured under different lighting conditions and crowd levels and also encompass a variety of screen sizes. Further, to improve the robustness, standard \emph{data augmentation} techniques (involving rotational and translational transformations to the training data) are used to synthetically enhance the training data corpus. 

Note that the Screen Extractor operates on each individual image (independent of whether the frame includes actual Manchester-encoded or transition frames), and is thus independent of screen refresh or camera frame rates. However, in our eventual mobile-based implementation, we shall see that we will prefer to run the Extractor \emph{intermittently}, on selected frames, to better balance the processing throughput and the screen extraction accuracy.

\vspace{-0.1in}
\subsection{DNN-Based Collective Bit Decoder} \label{subsec:decoder}

We next present a Convolutional Neural-Network (CNN) model that can extract the subtle temporal variation of Blue channel intensity across each individual cell, where each cell consists of a subset (e.g., $100 \times 100$) pixels. Under practical conditions, one can expect a screen extraction error of N pixels (N >= 0) in horizontal and/or vertical directions (e.g., a translation of 30 pixels in both directions). Clearly, the grid-based decoding techniques suffer from cross channel interference if $N > 0$. When $N \ge 0.3\times cell\_size$ (e.g., 30 pixels) in both directions, lower than 50\% of a transmitter cell is captured in the corresponding receiver cell (SNR < 0dB). To cope with these screen extraction errors, we propose to use a DNN model to infer input bits collectively based on the entire extracted screen. Figure~\ref{fig:cnn_model_features} shows our proposed DNN model structure which includes a 5-layer CNN and a fully connected layer. Our choice of a CNN-based approach is motivated by the remarkable capability of CNNs in vision-based pattern recognition tasks \cite{redmon2016you, howard2017mobilenets, girshick2015fast, szegedy2016rethinking}. In particular, CNN structure is well suited for this type of problem. A CNN practically comprises a large number of filters on the input images with different functionalities (e.g., one favours the bright areas, the other favours the darker areas). It extracts the essential ``features'', which are highly correlated to the ``bits'' values, through several filter layers. Essentially, each feature is computed from a receptive field which may easily cover many cells.

\begin{figure}[t]
	\centering
	\includegraphics[width=\columnwidth]{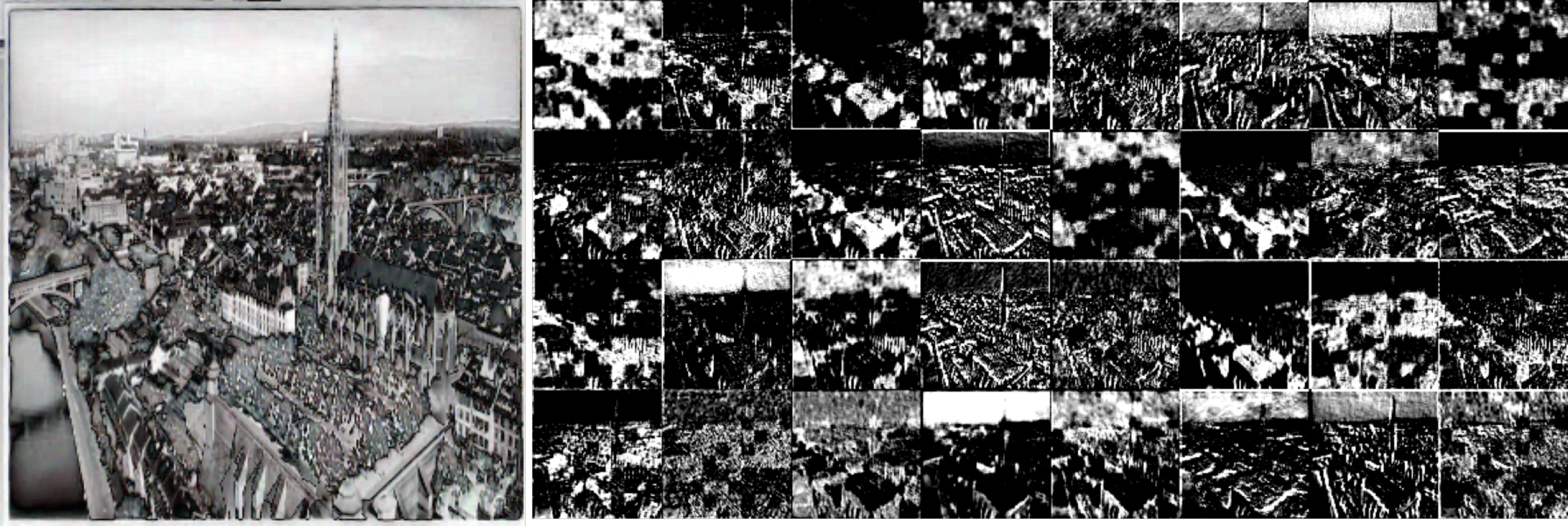}
	\vspace{-0.7cm}
	\caption{Left: input frame comprises the blue channel of the last 3 frames (with screen extraction error=($\approx 20\%$ of a cell). Right: Intermediate feature map at the 2nd convolutional layer, showing the outlines of the data grid. }
	\vspace{-0.5cm}
	\label{fig:inter_feature}
\end{figure}

The input to our CNN model is an image with $L=3$ channels, with each channel corresponding to the pixel value of the Blue channel of $L$ consecutive frames. These $L$ frames effectively capture both the rising (F+) and falling (F-) sequence of Manchester coded images. (While \emph{$L=3$ is used as an exemplar, based on the Nyquist assumption that the camera sampling rate is twice the display frame rate, we shall generalize this approach} in Section~\ref{sec:discussion}). The first layer is a $1 \times 1$ convolutional layer, that is used to extract different \emph{temporal} relationships, for each pixel, across the $L$ consecutive frames. Several additional  convolutional layers then extract higher-level and multi-scale features in the input data,  corresponding to a hybrid mix of spatial (across the neighboring pixels of a single image) and temporal (across the corresponding pixels of multiple frames) features. We also apply Batch Normalization \cite{ioffe2015batch} to improve the training of the convolutional layers. A finally fully-connected layer, with a Sigmoidal activation function, is used to regress the value of each bit, such that the final output of the CNN is a vector of $M\times N$ bits. We emphasize this key property essential for robustness: \textbf{each of the final $M\times N$ bits of output is derived from the holistic processing of $L$ \emph{entire} frames, rather than via isolated processing of a predesignated subset of pixels.}  As an illustration of this, Figure~\ref{fig:inter_feature} shows the output of the second convolutional layer in the decoder model, computed over an imperfectly-extracted representative screen image. We can see that the grid-like separation is already observable at this second layer, indicating that the DNN is able to perform decoding without explicit \emph{a-priori} knowledge of individual cell boundaries.

\subsubsection{Training the CNN}

To train \namef CNN-based decoder, we use a  binary cross-entropy (BCE) loss function \cite{BCE}, which attempts to maximize the number of correctly  matched bits—i.e., it tries to minimize the bit error rate (BER). The training dataset creation first involves the generation of display content (still images, video snapshots, movie content) video frames embedded with random bits using Manchester coding. To easily generate ground-truth labels that distinguish individual positive ($F+$) and negative ($F-$) Manchester-encoded frames pairs and that allows precise extraction of screen coordinates via color-based segmentation~\cite{szeliskibook}), we insert a dummy landmark frame (all pixels colored RED, $(R,G,B)=(255,0,0)$) after every 10 Manchester frame pairs ($(F+,F-)$). The frames are encoded into an uncompressed video format and displayed on a G-sync desktop monitor at 60 FPS (we choose this monitor and frame rate to avoid the frame tearing effect~\cite{tearingeffect}). The displayed encoded content is recorded using a mobile phone camera (iPhone X) at 120 FPS and at 720p HD resolution ($1280 \times 720$ image pixels). Each pair of Manchester frames maps to 4 distinct frames captured by the camera: a fully positive frame ($F_+$), a transitional frame (from positive to negative--called $F_+^t$), a fully negative frame ($F_-$), and a transitional frame from the negative frame to the next pair ($F_-^t$). We consider the Blue channel images of the 3 frames that include the Manchester pair ($\{F_+, F_+^t, F_-\}$) to represent 1 training image for the representative display content.  To make the \name CNN model robust to different screen errors, we train it with artificial randomized errors in the screen coordinates (relative to the ground truth screen coordinates), with errors within the range $\pm0.5 \times cell\-width$ on both the \emph{X} and \emph{Y} axes. Each extracted region of screen pixels with artificial errors is resized to produce a $299 \times 299$ pixel training image. Our final annotated dataset consists of 22,500 samples observed by a fixed/stationary camera and 25,200 samples captured by a hand-held camera. Overall, we observe a validation accuracy of $\approx 92.0\%$ (fixed) and $\approx 90.0\%$ (handheld) after 100 training epochs.

\subsubsection{Handling Larger Grid Sizes}
As mentioned previously, the \name decoder effectively uses a CNN-based regressor to simultaneously estimate the entire set of $M \times N$ bits encoded in a single Manchester image pair. A larger value of $M\times N$ (classifier outputs) increases the theoretical SCC throughput (no. of bits/image), but requires a more complex DNN model with higher training and runtime computational overhead. The canonical \name CNN decoder is built and trained for a grid size of ($M=$10 x $N=$10). Larger grid sizes can then be handled by simply duplicating the canonical model--e.g., if a \name encoder uses a $20 \times 20$ grid, the decoder applies the $10 \times 10$ CNN model four times, each on a quadrant of the input frame. However, a larger grid size means fewer pixels/cell. Consequently, the increased throughput (bits/frame) from larger grid sizes might be counteracted by higher decoding error-- we shall empirically study the consequent tradeoff between grid sizes and overall \emph{goodput} in Section~\ref{subsec:gridsize}.

\section{Prototype Implementation}
\label{sec:implementation}
\subsection{Bit Encoder}
To facilitate the easy experimentation with \names, we implement an offline encoding process (in Python), where the intensity-modulated video files are generated in advance and then played back on the display screen.  \emph{As will be explained shortly, the encoding process involves an initial pre-processing step where the data bits are grouped into blocks and then augmented by an error correcting code.} We use OpenCV to generate intensity modulated JPEG images (with data embedded) from original video frames and store the resulting video file in .AVI format. Unless otherwise mentioned, the source videos have an original frame rate of 24 to 30 FPS and an HD resolution ($1920 \times 1080$). We take the first 100 frames (approx. 3 secs) from each video to create a test set. After modulation, each encoded video clip has a frame rate of 60 FPS (pair of Manchester encoded frames).

\vspace{-0.1in}
\subsection{Bit Decoder}
\label{subsec:implementdecoder}

\name was implemented in two separate phases to support different facets of experimental studies:

\noindent{\bf Phase 1:} We conducted an offline study on the robustness of \name to support the rapid evaluation of a large amount of video data. As we use a diverse set of test videos (which were not provided during the model training phase), the resulted test data contains $\approx$2TB of high quality video. We first implemented \name in a desktop environment where the decoding operation was offloaded from the smartphone. In this study, we use an iPhone X (running iOS 13.3)  and a 64GB RAM desktop with a Core-i7-7700 CPU running at 3.6 GHz and a Nvidia GTX-1080Ti GPU card. The iPhone X operates its camera at 120 FPS, and captures full-HD quality videos (resolution=$1920 \times 1080$). Captured videos are stored as ``.MOV" files, and transmitted to the desktop computer for processing. 
    
\noindent{\bf Phase 2:} We implemented \name on a mobile phone to study how \name supports real-time processing when running \emph{entirely} on a mobile device. We use an iPhone 11 Pro (running iOS v13.6), with Apple A13 chipset and 4GB of RAM, for the mobile implementation of \names. We port our system to Objective-C and use CoreML [43] for the neural network models, using the same 32-bit floating point precision for both desktop and mobile versions. We shall analyze the processing time of each component in \names, and the resulting design choices (e.g., the periodic execution of the ScreenNet on a smaller set of image frames) to make such a real-time mobile implementation possible. 

\vspace{-0.1in}
\subsection{Implementing Error Correction}

To build a fully-functional \name decoder that supports reliable communication in spite of inevitable decoding errors, we integrate a Reed-Solomon (RS)~\cite{reed1960polynomial} error-correcting source code.
 Figure~\ref{fig:cnn_model_features} shows the overall frame structure using a representative $10 \times 10$ bit encoded frame. As RS code with N redundant symbols can correct a maximum of N/2 erroneous symbols, we augment it with a checksum code and a sequence number to detect frames that are falsely decoded by the RS decoder. These augmented fields (RS code, checksum and sequence number) can be adjusted without needing to retrain the core neural network. 

\vspace{1mm}\noindent{\bf Checksum:} The 5-bit BSD~\cite{bsdcsum} checksum is computed over the data bits and sequence number, and is used to identify any frame that the RS error correction fails to correct.

\vspace{1mm}\noindent{\bf Sequence number:} The sequence number is incremented on each data frame, and is used by the decoder to identify and eliminate duplicate (e.g., transition frames) and incorrectly-decoded frames. Given the frame rate ($F_{d})$ and the camera rate $F_{c}$, two consecutively numbered data frames should have a separation of at least $Sep= \lfloor \frac{2*F_c}{F_d}\rfloor$ image frames. The decoder logic eliminates potential false positives by verifying that two frames differing by $x$ in their sequence numbers are at least $x*Sep$ frames apart. We empirically found that this combination of mechanisms provides  highly robust decoding (all incorrectly decoded frames are identified and filtered out) as long as the percentage of redundant RS-code bits is $\ge 40$\%.

\section{Microbenchmark Evaluation} \label{sec:microbenchmark}

Our evaluation (performed with IRB approval from respective institutions) utilizes the following key performance metrics: 
   
\vspace{1mm}\noindent{\bf (1) Flicker perception}, defined using the subjective Mean-Opinion-Score (\textbf{MOS}) measure and quantitative Peak-Signal-to-Noise-Ratio (\textbf{PSNR}) metric, computed as $PSNR = 10 \times \log_{10}(R^{2}/MSE)$, where $R$ is the maximum feasible pixel value (for $n$ bit resolution of pixels $R=2^{n}-1$), and $MSE = \sum_{1,1}^{M,N}(I(m,n) - \hat{I}(m,n))^2/(M \times N)$, where $I(m,n)$ and $\hat{I}(m,n)$ are the original and modified pixel intensities respectively.

\vspace{1mm}\noindent{\bf (2) Frame Error Rate (FER)}, defined as the fraction of transmitted encoded data frames that are correctly decoded by the receiver. Given the use of RS decoding, a frame is either correct (if the RS code can correct the errors) or completely incorrect (even if some bits in the frame may be correct). We report FER rather than BER (bit-error-rate) as FER counts only recoverable data, while BER includes both recoverable and unrecoverable data.

\vspace{1mm}\noindent{\bf (3) Goodput (GP)}, defined as the  total number of \emph{data bits}, per second, correctly decoded by the \name decoder. $GP$ is computed as $GP = D \times F_{r} \times (1- FER)$, where $D$ is the useful data in each frame ($D= [(10\times10)\times (1 - RS\_ecc\_rate)]-10$ in our system), $Fr=0.5F_{d}$ is the data frame rate (0.5 scaling due to manchester coding), $F_{d}$ is the display frame rate, and $FER$ is the frame error rate.

\subsection{Flicker perception}
\label{subsec:flicker}

\begin{figure}[!t]
\centering
\includegraphics[width=0.48\columnwidth]{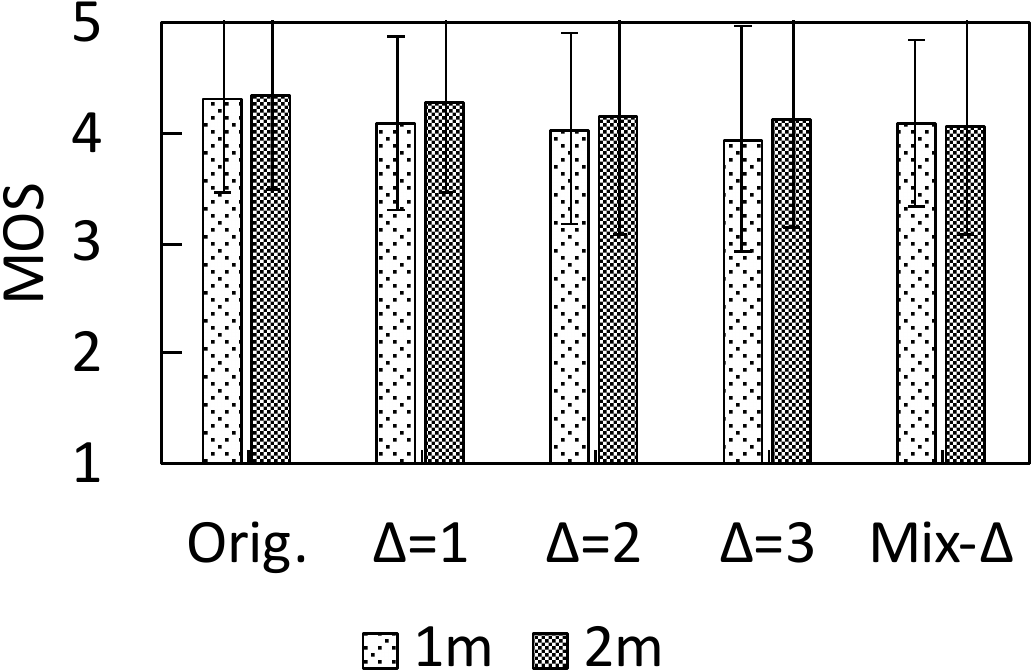}
\vspace{-0.3cm}
\includegraphics[width=0.48\columnwidth]{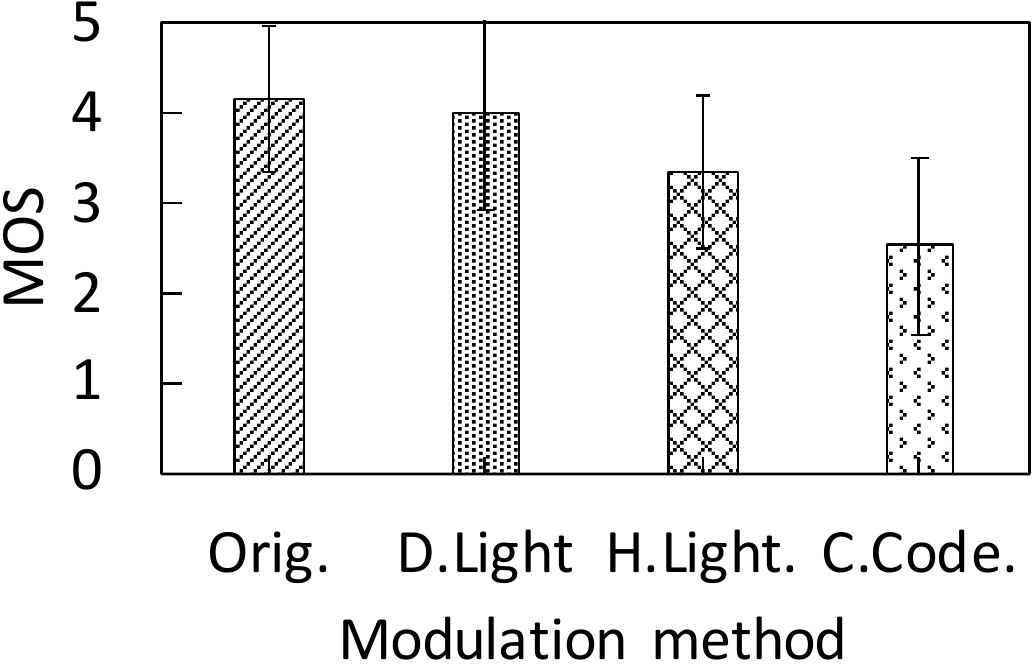}
\caption{User study MOS for: different $\Delta$ choices (left) and different encoding techniques (right).}
\vspace{-0.3cm}
\label{fig:mosresults}
\end{figure}

\begin{small}
\begin{table*}[th]
\begin{minipage}{0.49\linewidth}
    \centering
    \begin{tabular}{c|ccccc}
        \toprule
         & \multicolumn{5}{c}{Distance [m]}\\
         $\Delta$& 1.00 & 1.25 & 1.50 & 1.75 & 2.00\\
         \hline
         1 & 20.0/\bfblue{0.96} & 34.0/\bfblue{0.79} & 46.0/\bfblue{0.65} & 62.0/\bfblue{0.45} & 73.0/\bfblue{0.32}\\
         2 & 0.3/\bfblue{1.2} & 1.8/\bfblue{1.18} & 5.9/\bfblue{1.13} & 12.3/\bfblue{1.05} & 14.6/\bfblue{1.02}\\
         3 & 0.2/\bfblue{1.2} & 0.2/\bfblue{1.2} & 0.6/\bfblue{1.19} & 5.5/\bfblue{1.14} & 7.1/\bfblue{1.12}\\
         Mix & 0.4/\bfblue{1.2} & 0.9/\bfblue{1.19} & 2.1/\bfblue{1.18} & 7.6/\bfblue{1.11} & 11.7/\bfblue{1.06}\\
         \bottomrule
    \end{tabular}
    \caption{\name FER(\%)/\bfblue{GP(Kbps)} vs. ($d$, $\Delta$)}
    \vspace{-0.5cm}
    \label{tab:perf_delta}

\end{minipage}
\hfill
\begin{minipage}{0.49\linewidth}

    \centering
    \begin{tabular}{c|ccccc}
        \toprule
         & \multicolumn{5}{c}{Distance [m]}\\
         Rate& 1.00 & 1.25 & 1.50 & 1.75 & 2.00\\
         \hline
         0.6 & 0.2/\bfblue{0.90} & 0.3/\bfblue{0.90} & 1.0/\bfblue{0.89} & 4.9/\bfblue{0.86} & 7.2/\bfblue{0.84}\\
         0.5 & 0.4/\bfblue{1.2} & 0.9/\bfblue{1.19} & 2.1/\bfblue{1.18} & 7.6/\bfblue{1.11} & 11.7/\bfblue{1.06}\\
         0.4 & 0.7/\bfblue{1.49} & 2.1/\bfblue{1.47} & 6.5/\bfblue{1.4} & 12.6/\bfblue{1.31} & 18.4/\bfblue{1.22}\\
         0.3 & 4.0/\bfblue{1.73} & 4.5/\bfblue{1.72} & 12.9/\bfblue{1.57} & 22.2/\bfblue{1.4} & 28.7/\bfblue{1.29}\\
         \bottomrule
    \end{tabular}
    \caption{\name FER(\%)/\bfblue{GP(Kbps)} vs. ($d$, rate)}
  \vspace{-0.5cm}
    \label{tab:perf_rs}
\end{minipage}
\hfill
\end{table*}
\end{small}

\begin{small}
\begin{table*}[ht]
\begin{minipage}{0.45\linewidth}
    \centering
    \begin{tabular}{c|ccc}
        \toprule
         & \multicolumn{3}{c}{Distance [m]}\\
         Viewing Angle& 1.00 & 1.50 & 2.00\\
         \hline
         0\degree & 0.4/\bfblue{1.2} & 2.1/\bfblue{1.18} & 11.7/\bfblue{1.06}\\
         15\degree & 0.3/\bfblue{1.2} & 1.2/\bfblue{1.19} & 9.3/\bfblue{1.09}\\
         30\degree & 0.2/\bfblue{1.2} & 1.9/\bfblue{1.18} & 6.3/\bfblue{1.12}\\
         45\degree & 5.6/\bfblue{1.13} & 14.8/\bfblue{1.02} & 30.6/\bfblue{0.83}\\
         60\degree & 76.1/\bfblue{0.29} & 94.5/\bfblue{0.07} & 100.0/\bfblue{0.0}\\
         \bottomrule
    \end{tabular}
    \caption{\name FER(\%)/\bfblue{GP(Kbps)} vs. ($\theta,d$)}
     \vspace{-0.5cm}
    \label{tab:perf_angle}

\end{minipage}
\hfill
\begin{minipage}{0.45\linewidth}
    \centering
    \begin{tabular}{c|cccc}
        \toprule
         & \multicolumn{4}{c}{Screen detection error [cell size]}\\
         Error Type& 10\% & 20\% & 30\% & 40\%\\
         \hline
         SHF & 1.6/\bfblue{1.18} & 2.8/\bfblue{1.17} & 9.6/\bfblue{1.09} & 65.6/\bfblue{0.41}\\
         EXP & 1.2/\bfblue{1.19} & 1.6/\bfblue{1.18} & 2.4/\bfblue{1.17} & 5.9/\bfblue{1.13}\\
         SHR & 1.8/\bfblue{1.18} & 2.4/\bfblue{1.17} & 3.3/\bfblue{1.16} & 6.9/\bfblue{1.12}\\
         ROT & 1.6/\bfblue{1.18} & 1.9/\bfblue{1.18} & 2.8/\bfblue{1.17} & 5.7/\bfblue{1.13}\\
         \bottomrule
    \end{tabular}
    \caption{\name FER(\%)/\bfblue{GP(Kbps)} vs. Screen Extraction Error}
     \vspace{-0.5cm}
    \label{tab:perf_offset}
\end{minipage}
\hfill
\end{table*}
\end{small}

\begin{small}
    \begin{table}[t]
    \raggedleft{
    \begin{minipage}{0.37\linewidth}
        \begin{tabular}{c|c}
            \toprule
            Lighting& FER/GP  \\
            \hline
            eFL+BG&2.8/\bfblue{1.17}\\
            eFL+LED&1.4/\bfblue{1.18}\\
            iFL+LED&6.5/\bfblue{1.12}\\
            iFL&9.7/\bfblue{1.08}\\ 
            \bottomrule
        \end{tabular}
        \caption{\name FER(\%)/\bfblue{GP(Kbps)} vs. lighting types.}
        \vspace{-0.7cm}
        \label{tab:perf_ambient}
    \end{minipage}}
    \raggedright{
    \hspace{0.1cm}
    \begin{minipage}{0.59\linewidth}
        \vspace{-0.5cm}
        \includegraphics[width=\columnwidth]{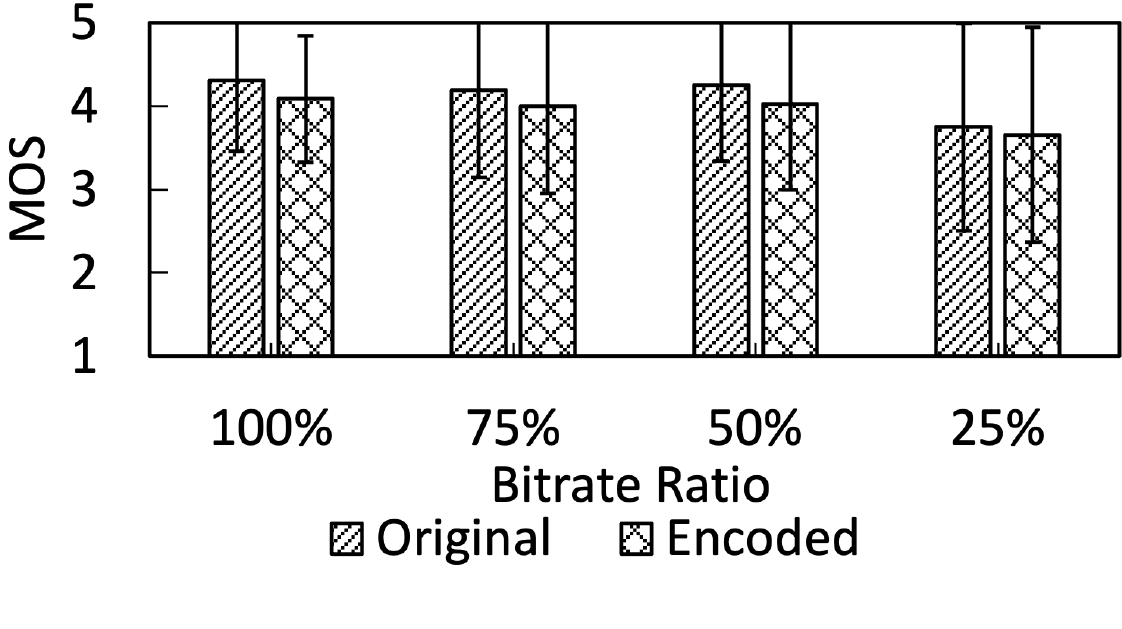}
        \vspace{-1cm}
        \captionof{figure}{Effects of compression on perceived quality.}
        \vspace{-0.7cm}
        \label{fig:moscompress}
    \end{minipage}}
    \end{table}
\end{small}

To evaluate the imperceptibility of \name encoding, we conducted user-studies where individual participants were each shown a total of 84 ten-second long video clips on a 25" display screen. The videos were randomized and included original videos with no encoding, encoded videos with different combinations of $\Delta$ (1, 2, 3 and a mix of 2 and 3), and videos encoded using Chromacode~\cite{zhang2019chromacode} and HiLight~\cite{li2015real} techniques (we implemented the encoder based on the description in the publication \cite{zhang2019chromacode,li2015real}). Users had no background knowledge of encoding and were asked to only view and rate each video clip. The experiments were conducted at distances 1m and 2m, at a reasonably comfortable (to user) frontal viewing angle, in an indoor office room environment with white ceiling lighting. The ratings use a  5-point MOS scale:\{(1): Very unpleasant; (2): It's bad;  (3): It could be better;  (4): It's good; (5): Cannot differentiate from the original video\}. The experiment involved 17 participants (10 males, 7 females), with age ranging [18,32] (median$=$24 yrs). One of the users has \emph{astigmatism}, 3 have farsightedness, 6 have no eye conditions, and the rest have shortsightedness.

\subsubsection{\names~Flicker perception Results}

Figure~\ref{fig:mosresults} (left) plots the average MOS values (and the std. deviation) for the original (unmodified) videos, as well as \names-encoded videos with $B$-channel intensity modulation values $\Delta=\{1,2,3\}$. We also experimented with a \emph{Mix-$\Delta$} strategy, where we use $\Delta=2$ if the avg. $B$-channel intensity is higher than a threshold (30), and $\Delta=3$ otherwise. We see that (a) the MOS scores stay almost unchanged (compared to the original) as long as $\Delta \le 2$, and (b) as expected, the flicker perception diminishes (MOS increases) when the viewing distance increases. The Mix-$\Delta$ strategy has the same perception score as a fixed value of $\Delta=2$. However, Mix-$\Delta$ was seen experimentally to  provide higher decoding accuracy, and is used as the \emph{default} encoder, unless otherwise specified.

\subsubsection{Comparative Evaluation of Flicker Perception}

From the MOS reported in Figure~\ref{fig:mosresults} (right), we can see that \namef score is significantly higher (avg=4), compared to both HiLight (avg=3.36) and Chromacode (avg=2.53), thereby validating \namef selective use of $B$-channel intensity modulation. The result is also validated by the higher average PSNR value (PSNR=42) for \names, compared to PSNR=41 for HiLight and PSNR=36 for ChromaCode. These results demonstrate that \name provides better viewer experience at typical display rates (60 FPS), compared to these prior state-of-the-art techniques.

\subsubsection{\name~Flicker Perception with Compressed Videos}

Highly compressed videos are relevant when a screen shows online videos. Consequently, the videos may be compressed at different ratios depending on the network condition. Note that \name~encodes the data after the video has been compressed, so the modulation amplitude ($\Delta$) is not affected by compression. In this experiment, we evaluate the perceived quality at 3 bitrate ratios including: $\{75\%, 50\%, 25\%\}\times original\_bitrate$. To avoid the bias caused by the lower quality of compressed videos, we measure MOS of both un-encoded and encoded videos for each compression ratios. Figure~\ref{fig:moscompress} shows that the MOS drop is fairly similar with un-compressed and moderately compressed videos (bitrate ratio $\ge50\%$). However, when the videos are highly compressed (bitrate ratio $=25\%$), the MOS drop is significantly lower and the standard deviation is considerably higher. This suggests that it is difficult to notice the flickers caused by \name~modulation among temporal and spatial artifacts caused by compression.

\subsection{FER and Goodput under different conditions.}
\label{subsec:micro}

We next evaluate the decoding performance of \name under a variety of instrumented settings. All the experiments were conducted in a meeting room which has a mix of fluorescent lamps and LEDs. We evaluate \name using 50 different short video clips (containing a mix of bright/dark content, slow/fast motion scenes, and low/high texture), downloaded from a video sharing website \cite{pexel} which provides free-of-use (commercial/non-commercial) video clips. Each result is obtained via 5 separate runs, each using 10 randomly selected clips. For these studies, the receiver camera is stationary (tripod-mounted), and the screen coordinates are precisely extracted via edge \& contour analysis in OpenCV using a special \emph{Red} frame displayed at the beginning of each experiment.

We evaluate \name with the default grid size $10 \times 10$. Each $10 \times 10$ data frame contains 40--70 bits of random data, 5 bits of sequence number, 5 bits of checksum, and 50--20 bits of RS error correction code with a symbol size of 5 bits. By default, we utilize 50 \emph{redundant} bits (50\% of a data frame), implying that the RS-Code can correct any data frame whose BER is $\le$5\% (5-bit symbols). We also evaluate \name with larger grid sizes: $\{20 \times 10, 20 \times 20, 30 \times 30\}$, with the percentage of 
redundancy bits being set to 60\%, 70\% and 80\%, respectively, thereby ensuring that it can correct frames with BER $\le$5\%.

\subsubsection{\name vs. $d$ (Viewing distance)}
In this experiment we place the camera at $0^\circ$ (perpendicular to the screen), and measure the FER for $d=$\{$1m$, $1.25m$, $1.5m$, $1.75m$, $2.00m$\} under \emph{perfect screen extraction}. Table~\ref{tab:perf_delta} lists the \emph{FER} and \emph{GP} values for a 10 X 10 data frame, with the RS-Code=50\%. We see that the Mix-$\Delta$ scheme simultaneously provides low values of FER (<0.15) and high goodput (>1 Kbps), even at larger viewing distances ($d=$2m). We also study the sensitivity of the decoding performance as the code rate (i.e., percentage of RS-coded redundancy bits), is varied over the set \{60\%, 50\%, 40\%, 30\%\}. Table~\ref{tab:perf_rs} lists the corresponding \emph{FER} and \emph{GP} values, as a function of $d$. We see that, as expected, a larger fraction of redundant bit results in a lower FER (with the FER remaining below 0.1 when the redundancy percentage is 50\% or higher). Conversely, a smaller fraction of redundancy bits (a higher code) rate allows \name to achieve goodput as high as 1.72 Kbps at shorter distance ($\le 1.25$meter) but results in a rapid goodput drop when the distances are larger ($GP=1.3$Kbps when $d=$2m). \emph{We infer that a larger redundancy percentage allows the goodput performance to remain stable across a wider operating range.}

\subsubsection{\name vs. $\theta$ (Viewing Angle)}
We next study the performance of \name under 9 different screen viewing angles $\theta$ : \{$\pm$60$^\circ$,$\pm$45$^\circ$,$\pm$30$^\circ$, $\pm$15$^\circ$, 0$^\circ$\}, for 3 different viewing distances $d=$\{1m, 1.5m, 2m\} and code rate (redundancy percentage) = 50\%.  Table~\ref{tab:perf_angle} lists the FER and goodput values as a function of $\theta$. We see that the FER stays low ($\le 10\%$) and goodput remains high ($\ge 1$Kbps) over at least a $60\deg$ viewing range ($\pm30\deg$), but degrades when the viewing angle gets more acute. To a large degree, this degradation may also be attributed to our use of a TN LCD panel, which  intrinsically has a narrower viewing angle than higher-end LCD screens (e.g, ones with IPS panel). 

\subsubsection{\name with Inaccurate Screen Extraction} 
\label{subsec:acc_scrdet_err}
We now study whether \name can \emph{support robust SCC even when the screen extraction is not completely accurate}. To perform this in a controlled manner, we conduct several distinct perturbations of the ground-truth coordinates of the screen (in the camera's pixel coordinates), including: (1) SHIFT: introduce translational error in the screen border along both X\& Y axes, (2) EXPAND: Move the screen borders outward (the estimated screen now includes background pixels), (3) SHRINK: Move the screen borders inward (the estimated screen now excludes some of the screen pixels), and (4) ROTATE: Rotate the screen borders in clockwise/counter-clockwise direction. \emph{The error magnitude is expressed in percentage of a cell size.} For example, at $d=1.5$m (viewing a 25" screen), a cell in a $10 \times 10$ grid has a size (in the camera image) of $\approx 55 \times 30$ pixels. Accordingly, a SHIFT of 30\% means introducing a translation error of 17 and 9 pixels in the vertical and horizontal borders, respectively. Table~\ref{tab:perf_offset} tabulates the resulting FER and GP. We see that \name is indeed \emph{robust to inaccurate screen extraction}, tolerating as much as 30\% noise in the screen extraction process without a dramatic increase in FER. SHIFT=40\% has a sharply higher FER, as the grids/cells on the screen boundary now lose over 64\% of their constituent pixels.

\subsubsection{\name vs. Different Ambient Lighting}
We next study the robustness of \name to different ambient lighting environments, both with and without outdoor ambient lighting (represented as BG). The \name Bit Decoder CNN was trained with the videos recorded under a combination of LED lights and fluorescent lamps (eFL) that use electronic ballast--this (eFL+LED) combination represents the \emph{default} ambient lighting. The eFL lamps have considerably lower flickers compared to older fluorescent lamps using inductive ballasts (iFL). We evaluate \name under a variety of additional lighting conditions, including (a) only iFL lights, (b) iFL + LED, and (c) with outdoor ambient lighting in the background (eFL+BG).  Table~\ref{tab:perf_ambient} plots the FER and GP values (for viewing distance $d=1.5$m) across these lighting alternatives. We see that \namef goodput remains steady across different ambient lighting choices. While the FER is higher when only iFL is used, we should note that iFL is an outdated technology that is being progressively phased out. 

\subsubsection{\name vs. Grid size} 
\label{subsec:gridsize}
\begin{figure}
	\centering
	\includegraphics[width=\columnwidth]{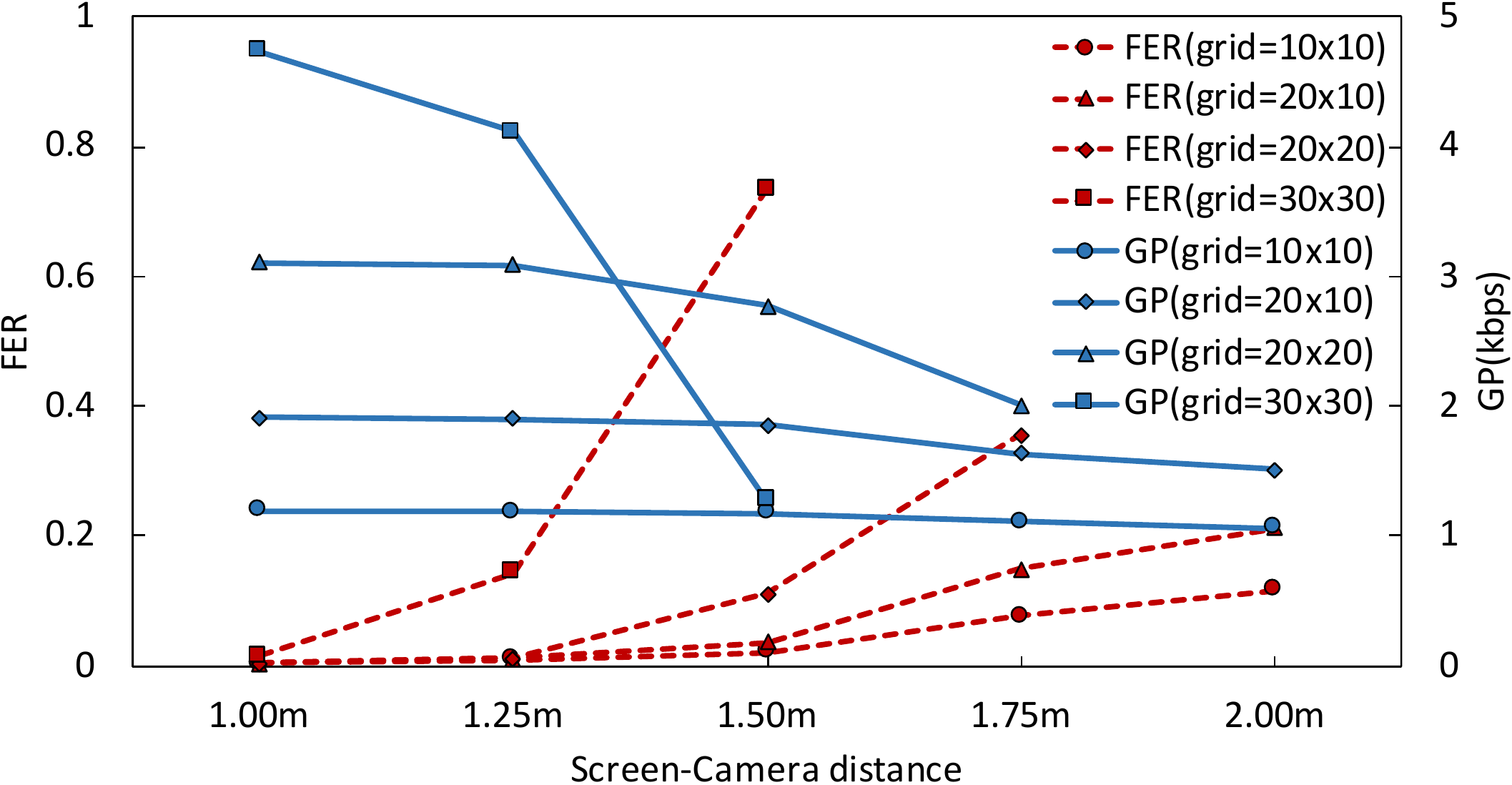}
	\vspace{-0.5cm}
	\captionof{figure}{\name FER and GP vs. $d$ at different grid sizes.}
	 \vspace{-0.2in}
	\label{fig:acc_dense}
\end{figure}

As discussed earlier in Section~\ref{sec:system}, \name supports larger grid size by repeating applying the $10 \times 10$ decoding pipeline on the input video frames. Of course, a larger grid size implies a smaller number of pixels/cell, and potentially higher decoding FER. Figure~\ref{fig:acc_dense} compares the FER and GP of different grid sizes at different distances. In comparison to the basic grid size of $10 \times 10$, the $20 \times 10$ grid size achieves comparable FER up to $D=1.5$m, but degrades more quickly at distances larger than 1.5m. In general, as $D$ increases, larger grid sizes cause the pixels/cell to be too low, leading to a sharp rise in FER. However, larger grids provide significantly higher goodput at short distances--e.g.,at $D=1.0$m, \name achieves a goodput of  4.7kbps using a $30 \times 30$ grid.

\noindent \textbf{Key Takeaways:} Overall, through these micro studies, we show that our DNN-based holistic decoding mechanism is robust, and not over-fitted to any specific content, screen border and aspect ratio. The test set is diverse, with 50 different clips of different contents, varying in different rates of dynamic change in screen content, and encompassing vistas ranging from nature to urban environments. \namef steady performance across different grid sizes also implicitly demonstrates that the \name is robust to different aspect-ratios. For example, for a 20x10 grid, the input frame is corresponding to an aspect-ratio of $\approx 10:12$, whereas the aspect ratio changes to $16:9$ in the case of a 10x10 grid. \name achieves similarly high accuracy across both grids at a distance $d=$1m; of course, the larger grids implicitly require shorter viewing distance  because of the smaller cell size.

\section{Evaluation of \name System}\label{sec:evaluation}
\subsection{\name Performance: User-held Smartphone}
\label{subsec:realworld}

\begin{figure}
\begin{minipage}{0.49\linewidth}
    \centering
	\includegraphics[width=\columnwidth]{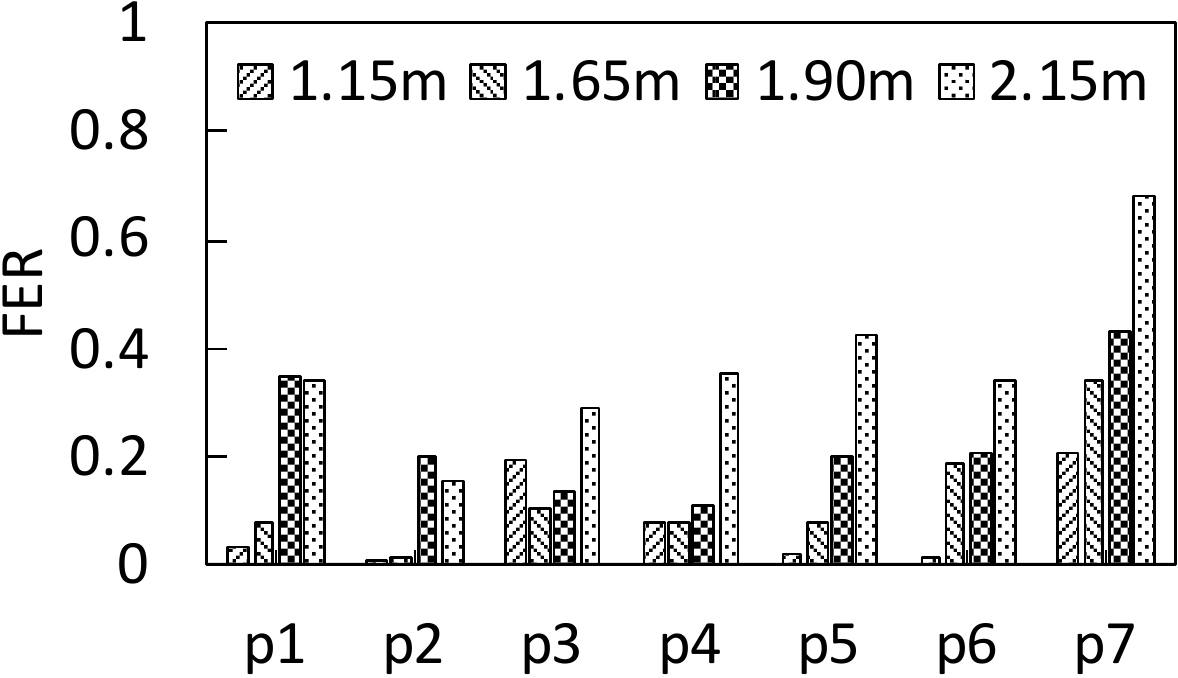}
\end{minipage}
\hfill
\begin{minipage}{0.49\linewidth}
    \centering
	\includegraphics[width=\columnwidth]{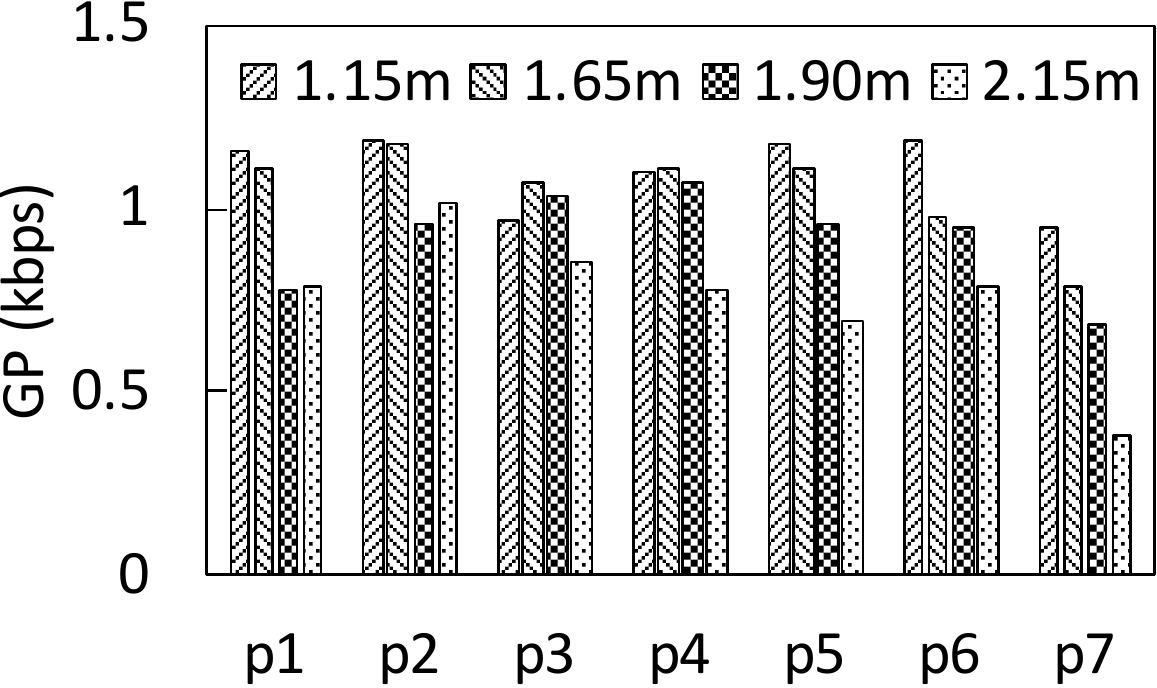}
\end{minipage}
\vspace{-0.2cm}
\caption{\name FER (left) and GP (right) across different distances under real world artefacts.}
\vspace{-0.3cm}
\label{fig:perf_handheld}
\end{figure}

We next study the real-world performance of the actual \name implementation. The goal of these experiments is to study the overall accuracy of the \emph{Screen Detector} and \emph{Bit Decoder} components, under typical artifacts of human smartphone usage. Accordingly, for these studies, the ML components are executed on the desktop computer. The studies involved 7 users using a smartphone (iPhone X) to record a display screen showing \name video clips. Each user sat on a chair, placed at multiple distances=\{1.15m, 1.65m, 1.90m, 2.15m\} from the screen, and recorded 10 randomly chosen (out of the 50 benchmarks) videos. To ensure motion-related artifacts, each user was instructed to \emph{not} rest their elbow on the chair arms. 

\subsubsection{\name Screen extraction accuracy}

We analyzed the performance of the \name Screen Extractor, using two widely-used metrics: (1) Intersection-over-union ($IoU$), defined as $IoU = (A_T \cap A_P)/(A_T \cup \ A_P)$, and (2) Intersection-over-Correct ($IoC$), defined as $IoC = (A_T \cap A_P)/(A_T)$, where $A_T$ represents the true screen coordinates and $A_P$ represents the screen coordinates estimated by the Screen Extractor. While a high $IoU$ is needed to reduce the amount of spurious background content in the extracted screen, a high $IoC$ is necessary to ensure that the \name decoder has access to the entire screen area. The optimal ($IoU, IoC$) combination is achieved by tuning the kernel size (see Table~(\ref{tab:iou_ioc})) of the dilation step in the screen extractor pipeline. By default, we use a kernel size of $2\times2$ (highlighted). We see that \name screen extractor is able to include 97\% of the screen under both indoor and outdoor conditions, and largely exclude the spurious background content (especially in indoor settings).

\begin{table}[ht]
	\centering
	\begin{tabular}{c|ccc}
		\toprule
	    & \multicolumn{3}{c}{Kernel size}\\
		& $1\times1$ & \cellcolor{lightgray}$2\times2$ & $3\times3$\\
		\hline
		Indoor & 0.93/\bfblue{0.95} & \cellcolor{lightgray}0.89/\bfblue{0.97} & 0.83/\bfblue{0.99}\\
		Outdoor & 0.83/\bfblue{0.97} & \cellcolor{lightgray}0.82/\bfblue{0.97} & 0.80/\bfblue{0.94}\\
		\bottomrule
	\end{tabular}
	\caption{Screen Extractor IoU/\bfblue{IoC}}
	\vspace{-0.25in}
	\label{tab:iou_ioc}
\end{table}

\subsubsection{\name FER and GP under real-world artefacts}
We evaluate the robustness of the performance of \name under real world human mobile device handling scenarios. In Figure \ref{fig:perf_handheld} we plot \namef FER and GP across the 7 users. In general, \names~ achieves relatively low FER $\le 0.2$ and high GP ($\ge 0.95kbps$) even for a relatively long viewing distance of 1.9 meters (for at least 5 of the 7 users). At $d=1.15$meter, the goodput is usually $\ge 1.1$Kbps. The poorer performance (much higher FER) of user P7 was due to the significantly higher amount of observed hand tremor, which causes significant image blur. As a point of contrast, note that, (a) even for a very short viewing distance ($d=0.7$meter), the default HiLight implementation  (Section~\ref{subsec:screenerror}) exhibits a much higher BER of $\approx 29$\% under a comparable screen detection accuracy (IoU= 89\%), while (b)  As reported in~\cite{zhang2019chromacode}, Chromacode achieves only $0.24$ Kbps at 1.8m ($0.47$ Kbps with a 120FPS display), that too using a fixed camera (no movement artifacts).  Overall, the results demonstrate that a smartphone-based implementation of \textbf{\name offers significantly better (than comparative techniques) and robust performance under real-world usage artifacts}. 

\subsection{Smartphone Implementation for Real-time Application}
\label{subsec:smartreal}

We now further consider the use of \name in a real-world application, where \name is implemented entirely on the smartphone (no edge offloading). To support \emph{real-time} decoding of camera-captured SCC content, \name must balance both the \emph{accuracy} and \emph{latency} metrics. 

To understand the latency characteristics, we first profile the per-frame processing time of each \name component in the system. There are three components that consume noticeable processing time: (1) Screen Detector (a UNet-based DNN), (2) Perspective warping (using OpenCV) that transforms the predicted screen area into a square shape, and (3) the Bit Decoder (a CNN). The latency of the RS Decoder is negligibly low ($\le 100 \mu s$ even on a mobile phone) and can be excluded from the latency profiling.

\subsubsection{Processing time profile:}

We measure the processing time of \name in two different environments: (1) a desktop computer, and (2) a mobile device (previously described in Section~\ref{subsec:implementdecoder}).  The desktop environment comprises  a core-i7 7700 CPU, an AORUS GeForce GTX 1080 Ti GPU card and 64GB of DDR4 DRAM. We port our system to C++ with Tensorflow C++ API to fairly compare with the mobile implementation. 

\begin{figure}[ht]
    \vspace{-0.15in}
    \begin{minipage}{0.49\linewidth}
	\centering
	\includegraphics[width=\columnwidth]{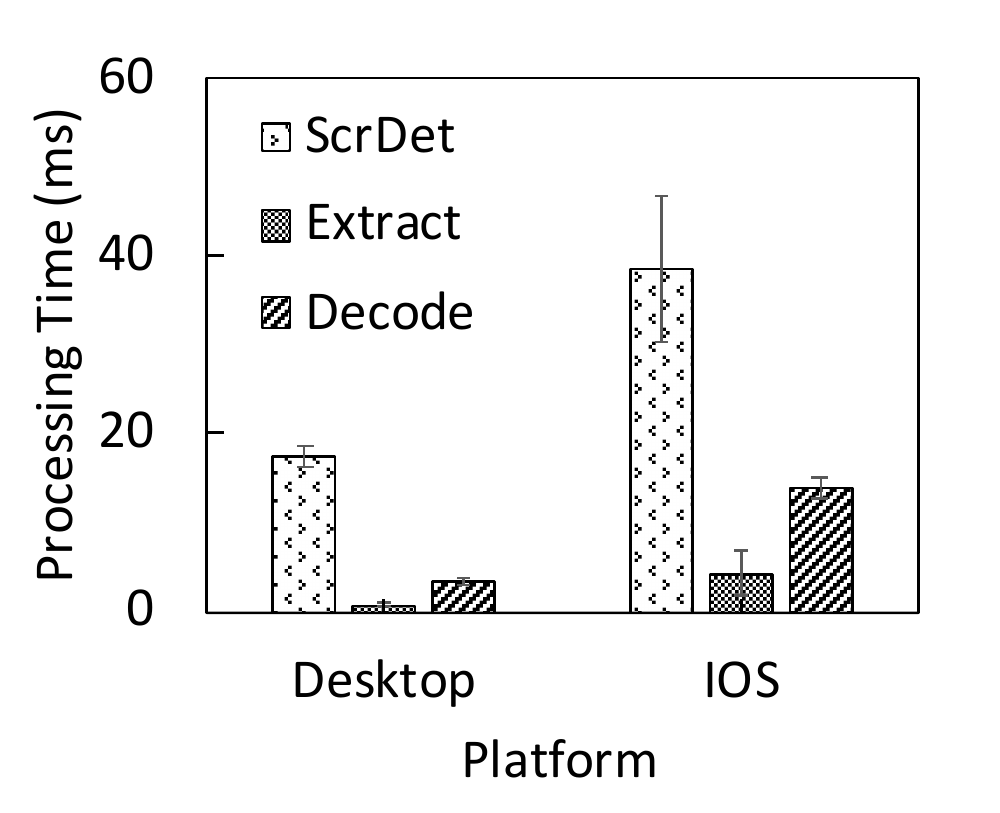}
	\end{minipage}
    \hfill
    \begin{minipage}{0.49\linewidth}
    \centering
	\includegraphics[width=\columnwidth]{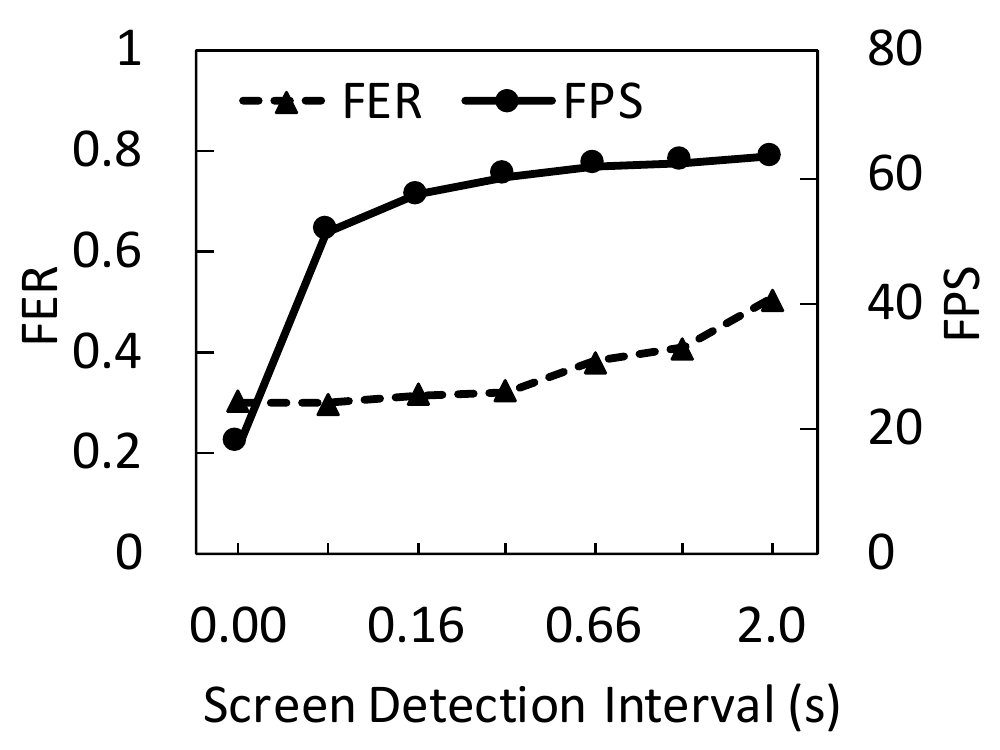}
    \end{minipage}
    \hfill
    \vspace{-3mm}
    \caption{\textit{LEFT}: Processing time of different \name Components (Desktop vs. iPhone 11Pro). \textit{RIGHT}: Accuracy of \name vs. Period between consecutive executions of screen detection, obtained when the smartphone is exhibiting motion perturbation while being held by a   \textit{standing} user.}
    \vspace{-0.1in}
    \label{fig:processing-time}
\end{figure}

Figure~\ref{fig:processing-time} shows the processing time of the three components in a desktop and a mobile device. The Screen Detector (ScrDet) consumes the highest processing time to the system with a value of 17.3ms on the desktop computer and 38.5ms on the iPhone. Similarly, the perspective warping and decoding take 0.75ms and 3.1ms on the desktop; and 4.1ms and 13.9ms on the iPhone accordingly. Given this total processing time of $\approx$ 56.5ms on the iPhone, \name can support a decoding rate of only $\approx$17.7FPS if all the components are executed on each image frame. However, we shall shortly show that it is not necessary to run the Screen Detector continuously.

\subsubsection{Supported FPS:}
\label{subsec:scrdet_interval}
To improve the supported FPS, our idea is to run the screen detector (the most time consuming component) only intermittently. This is based on the hypothesis that, for the common use case where a user stands or sits to watch the content, the screen position is unlikely to change significantly within a short time period (e.g. 100ms). To better understand this drift across frames, we study how much intermittent screen extraction affects FER with a \emph{standing} user to impose more hand motion artefacts. Figure~\ref{fig:processing-time}(right) shows that when the screen detector interval increases to 83ms, the achievable decoding rate almost triples to 51.4 FPS, with only a negligible increase in the FER. (To support a decoding rate of 60 FPs, the Screen Detector can execute once every 330ms, resulting in a relatively small  2.4\% increase in FER.) Beyond this point, the processing rate is bottle-necked by the latency of the Bit Decoder (which must, of course, be executed on each frame). Accordingly, \textbf{for our final real-time \name implementation, we execute the Screen Detector only once every 32 frames ($\approx$265msecs).}

\subsubsection{Energy consumption.}
\begin{figure}
	\centering
	\includegraphics[width=0.64\columnwidth]{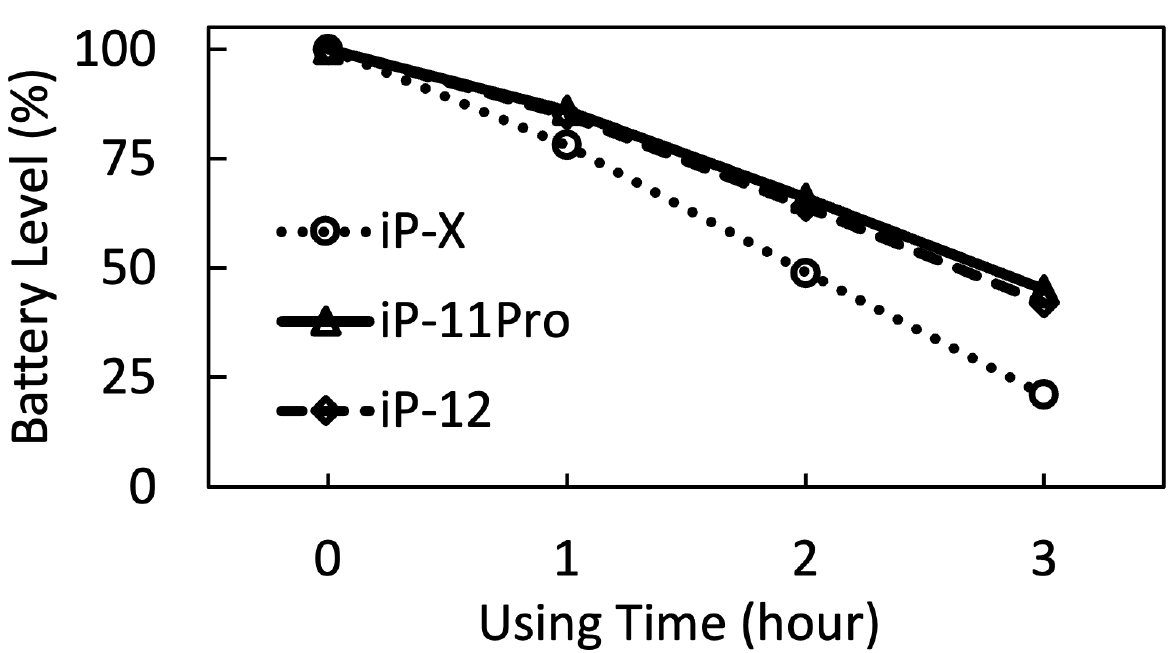}
	\vspace{-0.3cm}
	\captionof{figure}{Battery level drop while running \name continuously.}
	\vspace{-0.5cm}
	\label{fig:energy}
\end{figure}

To examine how long a mobile device can support \names, we let the smartphone run \name continuously for 3 hours and observe the battery level. We use three different smartphones for this experiment, including an iPhone X, an iPhone 11 Pro, and an iPhone 12. Figure~\ref{fig:energy} shows battery level of the three smartphones after every one hour. We can see that even the ``quite old'' smartphone (iPhone X) can support more than 3 hours of continuous processing, while the more modern ones consume lower than 60\% of battery capacity. We believe that an active operational lifetime of 4-5 hours (on a mobile device) is quite acceptable for our exemplar applications,  such as smart, multi-lingual subtitle/notification or subliminal audio ( Figure~\ref{fig:scc-usecase}), which the user is unlikely to use continuously for more than 1-2 hours.  

\begin{figure*}
\begin{minipage}{.48\textwidth}
 \centering
    \includegraphics[width=\columnwidth]{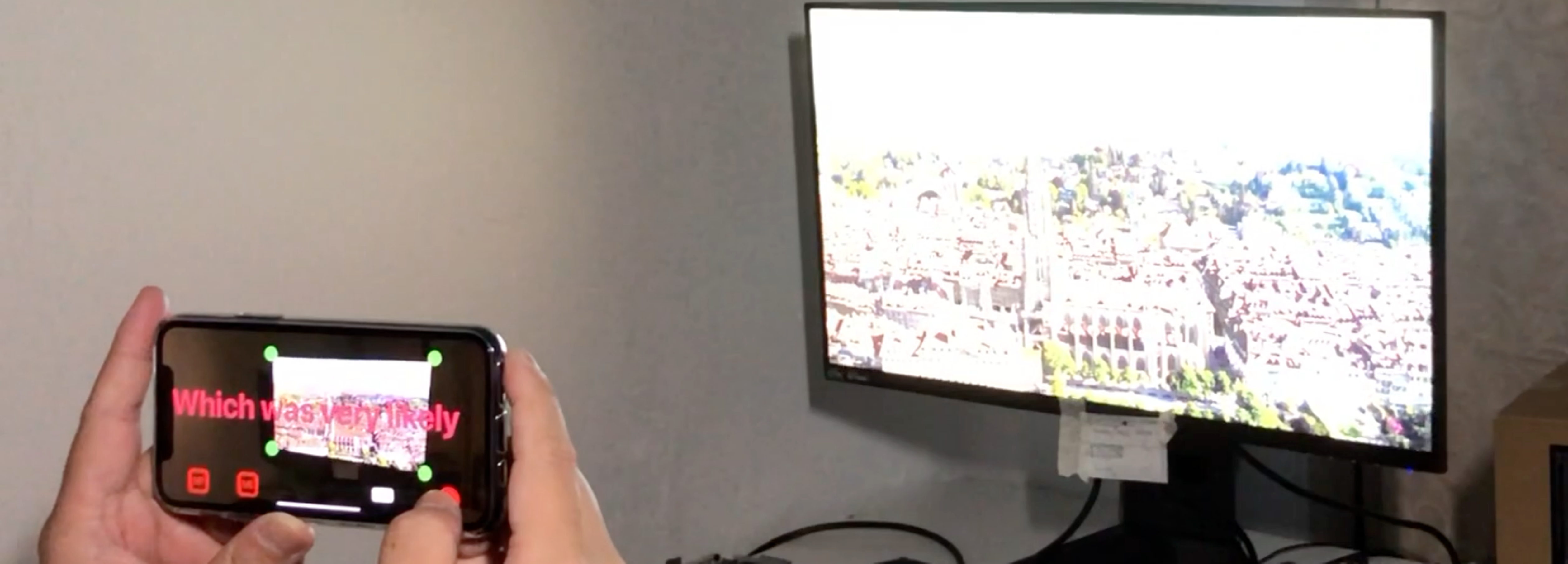}
    \vspace{-0.7cm}
    \caption{A smartphone, running \names-enabled captioning application, displays the text ( in `Alice in wonderland' textbook) decoded from a monitor.}
    \vspace{-0.3cm}
    \label{fig:mobile-app}
\end{minipage}%
\hfill
\begin{minipage}{.48\textwidth}
 \centering
 \includegraphics[width=\columnwidth]{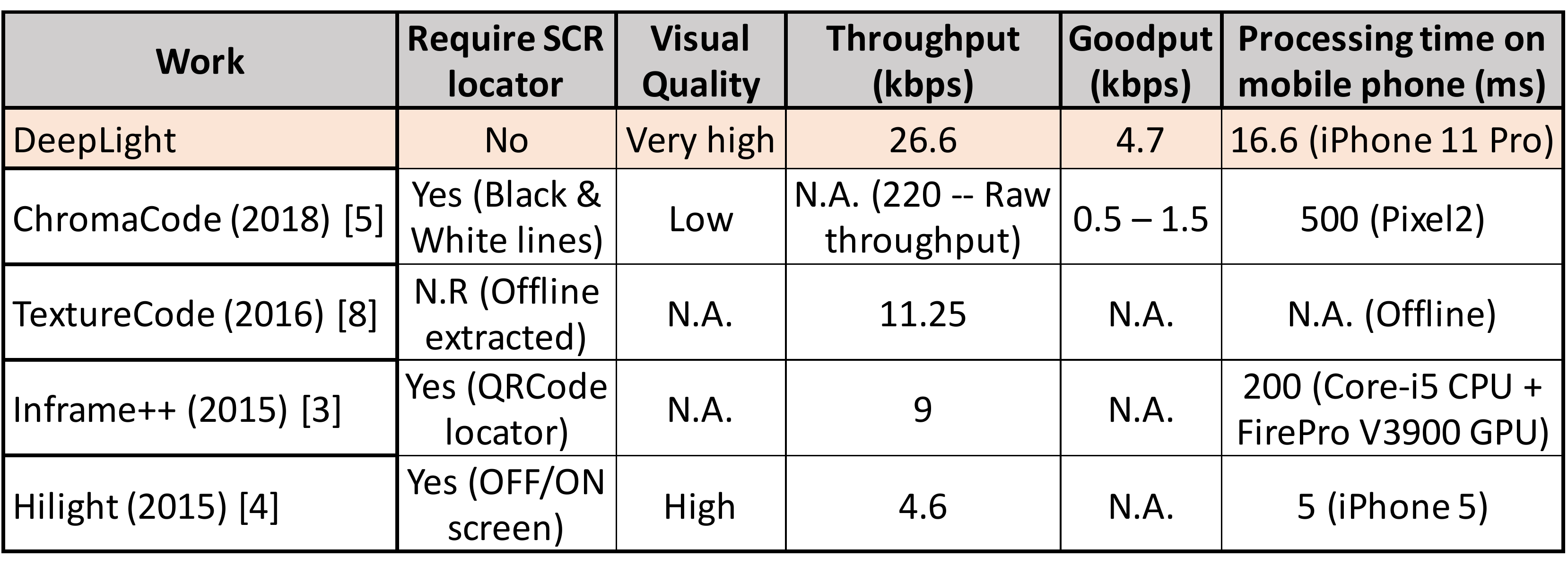}
 \vspace{-0.7cm}
 \caption{\name compared with prior works at 60FPS, 1m distance between screen and camera (N.A.= Not available).}
 \vspace{-0.3cm}
 \label{fig:comparison}
\end{minipage}%
\end{figure*}

\subsection{Closed-Captioning Application}
\label{subsec:closedcaption}

We implemented a live-captioning iOS-based application (on the iPhone 11 Pro we used in our studies). We first encode part of the text-book "Alice in Wonderland" into a video clip using blue-channel modulation as described in Section~\ref{sec:system}. On the mobile device side, we integrate \name into a self-developed camera recorder application which captures camera frames at 120FPS. To make sense of the response time of \names, our application works in a triggered manner. Whenever a user presses the ``START'' button, it executes the entire \name pipe-line to process the latest 32 frames (i.e., roughly 0.25 secs of video) stored in a circular-buffer. As discussed in Section~\ref{subsec:scrdet_interval}, \name only performs screen detection for the first frame (in 32 frames). Figure~\ref{fig:mobile-app} shows a representative photo of our live-captioning application running in front of a screen. The decoded text can be seen with the content of \emph{``Which was very likely"}. The estimated 4 corners of the screen are also shown as 4 ``green" dots on the mobile screen. Note that \name is able to decode the SCC data correctly, in spite of the modest errors in screen coordinate estimation.

\subsection{Comparison with prior works}

To compare \name with prior works, we first formalize 3 distinct technical terms: (a) \emph{Raw throughput} is defined as the total number of transmitted bits multiplied by the bit error rate; (b) \emph{Throughput} is defined as the total number of bits correctly received after RS-based error correction; (c) \emph{Goodput} is defined as the number of useful bits correctly received, excluding the overhead of error correction. \textit{Note that the raw throughput is not useful if the error rate is high}. For example, for a data transmission rate=2Mbps (1920*1080 resolution), the raw throughput would be 1Mbps if the receiver made a random guess; however, the actual number of decoded \emph{symbols} would be vanishingly low. As the source code of the state-of-the-art (Chromacode~\cite{zhang2019chromacode}) is not available, we compare \name performance with the reported performance of previous works. However, we implement the encoder of Chromacode~\cite{zhang2019chromacode} and HiLight~\cite{li2015real} to measure the perceived visual quality. To fairly compare different approaches, we normalized the throughput and goodput to a common screen frame rate of 60FPS, with a viewing distance of 1m. 

Figure~\ref{fig:comparison} provides a comparison of \name against other state-of-the-art approaches. Note that, due to our CNN-based decoding techniques, \name does not require either special display patterns or explicit ON/OFF activation of the screen to identify the screen coordinates. \namef use of Blue-channel modulation also offers significantly higher MOS (greater imperceptibility) compared to the alternatives. Due to the large number of videos assigned to each participant, we only evaluate the visual quality of \names, the most recent SCC work (Chromacode~ \cite{zhang2019chromacode}) and HiLight~\cite{li2015real} (which shown to achieve low flicker at 60FPS). In terms of throughput, TextureCode \cite{nguyen2016high} replicated Inframe++~\cite{wang2015inframepp} and HiLight~\cite{li2015real} and reported the corresponding throughput values (TextureCode definition of `correctly received bits' is equivalent to our Throughput definition). We borrow the throughput values reported in TextureCode \cite{nguyen2016high}, but normalized them to 60FPS instead of 120FPS. The real goodput values were not reported in those studies. Chromacode~\cite{zhang2019chromacode} reported raw throughput and goodput, but the throughput was not reported. We estimate the raw throughput and goodput of Chromacode from the reported values and normalize them to 60FPS. In general, \name outperforms prior works in terms of both throughput and goodput. At 1m, using a grid size of $30\times30$, \name achieve a throughput and goodput of 26.6Kbps and 4.7Kbps respectively. In terms of processing time, \name achieves an average full-pipeline processing time of $\approx 16.6ms$; in contrast, HiLight~\cite{li2015real} achieves a lower processing time of $\approx 5ms$. Though \name does not achieve the lowest processing time, it supports $\approx60FPS$ on a mobile platform (iPhone 11 Pro).

\section{Discussion}\label{sec:discussion}
In this section, we discuss some of the limitations of our current work and possibilities for future improvement.

\vspace{1mm}\noindent \textbf{Grid Sizes and Smarter Spatial Codec (coding \& decoding):}
While \name currently assumes explicit knowledge of the grid size, we believe that this restriction can be overcome by having a preamble frame (with a predefined grid size) bearing metadata (e.g., grid size) of payload frames. In addition, it might be possible to design a smart encoder-decoder co-design that learns to optimize both shape and intensity modulation (e.g., using Generative Adversarial Network) for both low flicker and high goodput. In long term, it is intriguing to explore the possibility of a single model that performs both screen extracting and decoding.

\vspace{1mm}\noindent \textbf{Handling Moving Users: } 
Current \name implementation does not work well if the user is walking while holding the smartphone due to (a) the significant change in screen position across even consecutive frames caused by user's steps, and (b) the blurry nature of such in-motion images. While the first problem may be addressed by introducing a smaller ScreenNet model suitable for continuous per-frame execution, the second can be addressed by well-known camera stabilization techniques \cite{baluja2017hiding, cardani2006optical, rahmati2009noshake}.

\vspace{1mm}\noindent \textbf{Supporting Different Display Rates:} 
We additionally experimented with a 30FPS video on our default 25" screen ($F_D=60Hz$); \name was seen to achieve high decoding accuracy ($FER=0.03$, GP=$0.58$Kbps, $d=1$m). Formally speaking, a Manchester pair of 30FPS video is translated into 8 camera frames ($F_C=120$FPS), in which the transition from F+ to $F-$ (or vice versa) still occurs across a $L=3$ ($F+,T,F-$) frame triple, but albeit with this frame triple occurring less frequently. After finding the temporal separation between such frame triples, \name can compute the actual frame rate.

\section{Conclusion}\label{sec:conclusion}
In this paper, we presented \names, an end-to-end SCC system whose receiver uses two novel DNN models to both (a) perform reasonably accurate extraction of screen content from a captured image without any external knowledge, and (b) decode the encoded bits collectively, without explicit partitioning of the screen content into constituent cells. These two innovations help us develop a practical SCC system that can work in the real-world---
without special frame markers, under different lighting conditions and with varying viewing distances and angles. \name consistently supports data goodput of $ 1$Kbps at viewing distance of up to $\approx 2$m, with a viewing angle of more than $\pm30^\circ$. These innovations allow \name to support several new classes of in-the-wild SCC applications via \emph{personal mobile/wearable devices}. These advances are also testimony to the power of carefully introducing ML pipelines in advanced communication systems.

\section{Acknowledgments}\label{sec:acknowledgements}
This work was supported by the National Research Foundation, Singapore under its NRF Investigatorship grant (NRF-NRFI05-2019-0007). Any opinions, findings and conclusions or recommendations expressed in this material are those of the author(s) and do not reflect the views of National Research Foundation, Singapore.







\balance
\end{document}